\documentclass[11pt, a4paper]{article}% Math and Physical Sciences Reference Style

\usepackage{geometry}

\usepackage{graphicx}
\usepackage{hyperref}
\usepackage{amsmath, amssymb, amsthm}
\usepackage{booktabs} % For formal tables
\usepackage{xspace} % For controlling spaces after newcommand
\usepackage[ruled,vlined]{algorithm2e} % Used for algorithms
\usepackage{mathtools}
\usepackage{balance}
\usepackage{enumitem}

\usepackage{proof-at-the-end}

\newcommand{\oea}{${(1+1)}$~EA\xspace}
\newcommand{\ocl}{${(1,\texorpdfstring{\lambda}{λ})}$~EA\xspace}

\newcommand{\mcl}{${(\mu , \lambda)}$~EA\xspace}
\newcommand{\opl}{${(1+\texorpdfstring{\lambda}{λ})}$~EA\xspace}
\newcommand{\saocl}{self-adjusting $(1,\texorpdfstring{\lambda}{λ})$~EA\xspace}
\newcommand{\saopl}{$(1+\{F^{1/s}\texorpdfstring{\lambda}{λ}, \texorpdfstring{\lambda}{λ}/F\})$~EA\xspace}
\newcommand{\ga}{${(1 +(\texorpdfstring{\lambda}{λ},\texorpdfstring{\lambda}{λ}))}$~GA\xspace}

\newcommand{\round}[1]{\ensuremath{\lfloor#1\rceil}}
\newcommand{\ones}[1]{\left\lvert#1\right\rvert_1}
\newcommand{\zeros}[1]{\left\lvert#1\right\rvert_0}

\newcommand{\onemax}{\textsc{OneMax}\xspace}
\newcommand{\ridge}{\textsc{Ridge}\xspace}
\newcommand{\jump}{\textsc{Jump}\xspace}
\newcommand{\JUMPK}{\ensuremath{\textsc{Jump}_k}\xspace}
\newcommand{\jumpk}{\JUMPK}
\newcommand{\cliff}{\textsc{Cliff}\xspace}
\newcommand{\CLIFFD}{\ensuremath{\textsc{Cliff}_d}\xspace}
\newcommand{\cliffd}{\CLIFFD}

\newcommand{\TWOMAX}{\textsc{TwoMax}\xspace}
\newcommand{\twomax}{\TWOMAX}
\newcommand{\ZEROMAX}{\textsc{ZeroMax}\xspace}
\newcommand{\zeromax}{\ZEROMAX}

\newcommand{\lambdainit}{\lambda_0}

\newcommand{\pimp}{p_{i,\lambda}^+}
\newcommand{\peq}{p_{i,\lambda}^0}
\newcommand{\ploss}{p_{i,\lambda}^-}

\newcommand{\Deltaloss}{\Delta_{i,\lambda}^-}
\newcommand{\Deltagain}{\Delta_{i,\lambda}^+}

\newcommand{\Deltah}{\Delta_{g_2}}

\newcommand{\E}[1]{\text{E}\left(#1\right)}
\newcommand{\Prob}[1]{\mathrm{Pr}\left(#1\right)}
\newcommand{\prob}[1]{\Prob{#1}}

\newcommand{\N}{\mathbb{N}}

\renewcommand{\epsilon}{\varepsilon}

\newcommand{\lambdabar}{\overline{\lambda_b}}

\usepackage{xcolor}
\newtheorem{theorem}{Theorem}[section]% meant for sectionwise numbers
\newtheorem{lemma}[theorem]{Lemma}
\newtheorem{corollary}[theorem]{Corollary}
\theoremstyle{definition}%
\newtheorem{definition}[theorem]{Definition}%

\title{Self-Adjusting Population Sizes for Non-Elitist Evolutionary Algorithms\\
Why Success Rates Matter}

\author{Mario Alejandro Hevia Fajardo\\
University of Sheffield\\
Sheffield, UK
\and
Dirk Sudholt\\
University of Passau\\
Passau, Germany}

\begin{document}

\maketitle

\abstract{Evolutionary algorithms (EAs) are general-purpose optimisers that come with several parameters like the sizes of parent and offspring populations or the mutation rate. It is well known that the performance of EAs may depend drastically on these parameters. Recent theoretical studies have shown that self-adjusting parameter control mechanisms that tune parameters during the algorithm run can provably outperform the best static parameters in EAs on discrete problems. However, the majority of these studies concerned elitist EAs and we do not have a clear answer on whether the same mechanisms can be applied for non-elitist EAs.
% Recent theoretical studies have shown that self-adjusting mechanisms can provably outperform the best static parameters in evolutionary algorithms on discrete problems. However, the majority of these studies concerned elitist algorithms and we do not have a clear answer on whether the same mechanisms can be applied for non-elitist algorithms.

We study one of the best-known parameter control mechanisms, the one-fifth success rule, to control the offspring population size~$\lambda$ in the non-elitist \ocl. It is known that the \ocl has a sharp threshold with respect to the choice of~$\lambda$ where the expected runtime on the benchmark function \onemax changes from polynomial to exponential time. Hence, it is not clear whether parameter control mechanisms are able to find and maintain suitable values of~$\lambda$.

For \onemax we show that the answer crucially depends on the success rate~$s$ (i.\,e.\ a one-$(s+1)$-th success rule). We prove that, if the success rate is appropriately small, the self-adjusting \ocl optimises \onemax in $O(n)$ expected generations and $O(n \log n)$ expected evaluations, the best possible runtime for any unary unbiased black-box algorithm. A small success rate is crucial: we also show that if the success rate is too large, the algorithm has an exponential runtime on \onemax and other functions with similar characteristics.}

\section{Introduction}
\label{intro}

Evolutionary algorithms (EAs) are general-purpose randomised search heuristics inspired by biological evolution that have been successfully applied to solve a wide range of optimisation problems. The main idea is to maintain a population (multiset) of candidate solutions (also called \emph{search points} or \emph{individuals}) and to create new search points (called \emph{offspring}), from applying genetic operators such as \emph{mutation} (making small changes to a \emph{parent} search point) and/or \emph{recombination} (combining features of two or more parents). A process of \emph{selection} is then applied to form the next generation's population. This process is iterated over many generations in the hope that the search space is explored and high-fitness search points emerge.

Thanks to their generality, evolutionary algorithms are especially helpful when the problem in hand is not well-known or when the underlying fitness landscape can only be queried through fitness function evaluations (black-box optimisation)~\cite{Eiben2015}.
Frequently in real-world applications the fitness function evaluations are costly, therefore there is a large interest in reducing the number of fitness function evaluations needed to optimise a function, also called optimisation time or runtime~\cite{BookNeuWit,Jansen2013,Auger2011,doerr-neumann-book}.

EAs come with a range of parameters, such as the size of the parent population, the size of the offspring population or the mutation rate.
It is well known that the optimisation time of an evolutionary algorithm may depend drastically and often unpredictably on their parameters and the problem in hand~\cite{LoboLimaMichalewicz2007,DoerrSurvey2020}. Hence, parameter selection is an important and growing field of study.

One approach for parameter selection is to theoretically analyse the optimisation time (runtime analysis) of evolutionary algorithms to understand how different parameter settings affect their performance on different parameter landscapes. This approach has given us a better understanding of how to properly set the parameters of evolutionary algorithms. In addition, owing to runtime analysis we also know that during the optimisation process the optimal parameter values may change, making any static parameter choice have sub-optimal performance \cite{DoerrSurvey2020}. Therefore, it is natural to also analyse parameter settings that are able to change throughout the run. These mechanisms are called parameter control mechanisms.

Parameter control mechanisms aim to identify parameter values that are optimal for the current state of the optimisation process. In continuous optimisation, parameter control is indispensable to ensure convergence to the optimum, therefore, non-static parameter choices have been standard for several decades. In sharp contrast to this, in the discrete domain parameter control has only become more common in recent years. This is in part owing to theoretical studies demonstrating that fitness-dependent parameter control mechanisms can provably outperform the best static parameter settings \cite{Badkobeh2014,Boettcher2010,Doerr2015,Doerr20201}. Despite the proven advantages, fitness-dependent mechanisms have an important drawback: to have an optimal performance they generally need to be tailored to a specific problem, which needs a substantial knowledge of the problem in hand~\cite{DoerrSurvey2020}.

To overcome this constraint, several parameter control mechanisms have been proposed that update the parameters in a \emph{self-adjusting} manner. The idea is to adapt parameters based on information gathered during the run, for instance whether a generation has led to an improvement in the best fitness (called a \emph{success}) or not. Theoretical studies have proven that in spite of their simplicity, these mechanisms are able to use \emph{good} parameter values throughout the optimisation, obtaining the same or better performance than any static parameter choice.

There is a growing body of research in this rapidly emerging area. Lässig and Sudholt~\cite{Lassig2011} presented self-adjusting schemes for choosing the offspring population size in (1+$\lambda$)~EAs and the number of islands in an island model. Mambrini and Sudholt~\cite{Mambrini2015} adapted the migration interval in island models and showed that adaptation can reduce the communication effort beyond the best possible fixed parameter. Doerr and Doerr~\cite{Doerr2018} proposed a self-adjusting mechanism in the \ga based on the one-fifth rule and proved that it optimises the well known benchmark function \onemax$(x) = \sum_{i=1}^n x_i$ (counting the number of ones in a bit string $x \in \{0, 1\}^n$ of length~$n$) in $O(n)$ expected evaluations, being the fastest known unbiased genetic algorithm on \onemax. Hevia Fajardo and Sudholt~\cite{Hevia2020} studied modifications to the self-adjusting mechanism in the \ga on \jump functions, showing that they can perform nearly as well as the \oea with the optimal mutation rate. Doerr, Doerr, and Kötzing~\cite{DoerrDK2018} presented a success-based choice of the mutation strength for a randomised local search (RLS) variant, proving that it is very efficient for a generalisation of the \onemax problem to a larger alphabet than $\{0, 1\}$. Doerr, Gießen, Witt, and Yang \cite{Doerr2019opl} showed that a success-based parameter control mechanism is able to identify and track the optimal mutation rate in the (1+$\lambda$)~EA on \onemax, matching the performance of the best known fitness-dependent parameter~\cite{Badkobeh2014}.
Doerr and Doerr give a comprehensive survey
of theoretical results~\cite{DoerrSurvey2020}.

Most theoretical analyses of parameter control mechanisms focus on so-called \emph{elitist EAs} that always reject worsening moves (with notable exceptions that study self-adaptive mutation rates in the \ocl \cite{DoerrAdaptive21} and the \mcl \cite{Case2020}, and hyper-heuristics that choose between elitist and non-elitist selection mechanisms~\cite{LissovoiMultimodal2019}). The performance of parameter control mechanisms in non-elitist algorithms is not well understood, despite the fact that non-elitist EAs are often better at escaping from local optima~\cite{Jagerskupper2007a} and are often applied in practice. There are many applications of non-elitist evolutionary algorithms for which an improved theoretical understanding of parameter control mechanisms could bring performance improvements matching or exceeding the ones seen for elitist algorithms.

We consider the \ocl on \onemax that in every generation creates $\lambda$ offspring and selects the best one for survival. Rowe and Sudholt~\cite{Rowe2014} have shown that there is a sharp threshold at $\lambda=\log_{\frac{e}{e-1}} n$ between exponential and polynomial runtimes on \onemax. A value $\lambda \ge \log_{\frac{e}{e-1}} n$ ensures that the offspring population size is sufficiently large to ensure a positive \emph{drift} (expected progress) towards the optimum even on the most challenging fitness levels. For easier fitness levels, smaller values of $\lambda$ are sufficient.

This is a challenging scenario for self-adjusting the offspring population size~$\lambda$ since too small values of $\lambda$ can easily make the algorithm decrease its current fitness, moving away from the optimum. For static values of ${\lambda \le (1-\varepsilon) \log_{\frac{e}{e-1}} n}$, for any constant $\varepsilon > 0$, we know that the optimisation time is exponential with high probability~\cite{Rowe2014}. Moreover, too large values for $\lambda$ can waste function evaluations and blow up the optimisation time.

We consider a self-adjusting version of the \ocl that uses a
success-based rule. Following the naming convention from \cite{DoerrSurvey2020} the algorithm is called self-adjusting $(1,\{F^{1/s}\lambda, \lambda/F \})$~EA (\saocl). For an update strength $F$ and a success rate $s$, in a generation where no improvement in fitness is found, $\lambda$ is increased by a factor of $F^{1/s}$ and in a successful generation, $\lambda$ is divided by a factor~$F$. If one out of $s+1$ generations is successful, the value of $\lambda$ is maintained.
The case $s=4$ is the famous one-fifth success rule \cite{Rechenberg1973,KMH04}.

We ask whether the \saocl is able to find and maintain suitable parameter values of $\lambda$ throughout the run, despite the lack of elitism and without knowledge of the problem in hand.

We answer this question in the affirmative if the success rate $s$ is chosen correctly. We show in Section~\ref{sec:poly-runtime} that, if $s$ is a constant with $0 < s < 1$, then the \saocl optimises \onemax in $O(n)$ expected generations and $O(n \log n)$ expected fitness evaluations. The bound on evaluations is optimal for all unary unbiased black-box algorithms~\cite{Lehre2012,Doerr20201}. However, if $s$ is a sufficiently large constant, $s \ge 18$, the runtime on \onemax
%(and other unimodal and multimodal functions with similar characteristics)
becomes exponential with overwhelming probability (see Section~\ref{sec:exponential-runtime}). The reason is that then unsuccessful generations increase $\lambda$ only slowly, whereas successful generations decrease $\lambda$ significantly. This effect is more pronounced during early stages of a run when the current search point is still far away from the optimum and  successful generations are common. We show that then the algorithm gets stuck in a non-stable equilibrium with small $\lambda$-values and frequent fallbacks (fitness decreases) at a linear Hamming distance to the optimum. This effect is not limited to \onemax; we show that this negative result easily translates to other functions for which it is easy to find improvements during early stages of a run.

To bound the expected number of generations for small success rates on \onemax, we apply drift analysis to a potential function that trades off increases in fitness against a penalty term for small $\lambda$-values. In generations where the fitness decreases, $\lambda$ increases and the penalty is reduced, allowing us to show a positive drift in the potential for all fitness levels and all $\lambda$.

To bound the expected number of evaluations, we provide two different analyses. In Section~\ref{sec:elitist-saopl} we refine and generalise the amortised analysis from~\cite{Lassig2011} to work with arbitrary hyperparameters $s > 0$ and $F > 1$. The main idea is to show that the number of evaluations is asymptotically bounded by the expected number of evaluations \emph{in unsuccessful generations}, denoted as $U_i$ for a fitness level~$i$. The latter is bounded by exploiting that when the offspring population size is large when entering a new fitness level, the generation is successful with high probability and then $U_i = 0$. This helps to show that $\E{U_i} = O(1/p_{i, 1}^+)$ where $p_{i, 1}^+$ is the probability of an improving mutation from fitness level~$i$.
The major downside of this approach is that it only works for elitist algorithms, i.\,e.\ an elitist \saopl.

In Section~\ref{sec:evaluations} we then present a different approach that does apply to the \saocl as well. We use the potential to bound the expected number of evaluations to increase the best-so-far fitness by $\log n$, reaching a new fitness value denoted by~$b$. The time until this happens is called an \emph{epoch}. During an epoch, the number of evaluations is bounded by arguing that $\lambda$ is unlikely to increase much beyond a threshold value of $O(1/p_{b-1, 1}^+)$, where $p_{b-1, 1}^+$ is the worst-case improvement probability as long as no fitness of at least $b$ is reached. Since at the start of an epoch the initial value of $\lambda$ is not known, we provide a tail bound showing that $\lambda$ is unlikely to attain excessively large values and hence any unknown values of $\lambda$ contribute to a total of $O(n \log n)$ expected evaluations.

% construct a novel  ``ratchet argument'': we show that, even when the fitness decreases, it does not decrease much below the best fitness seen so far. More precisely, with high probability, if $f(x_t)$ is the current fitness at time~$t$ and $f_t^* = \max\{f(x_{t'}) \mid t' \le t\}$ is the best fitness seen so far, then, with high probability, $f(x_t) \ge f_t^* - r \log n$ for an appropriate constant~$r$. Then we show that there is a constant probability that the best-so far fitness is increased by $\log n$ in a sequence of generations without fallbacks.
% We are hopeful that these arguments will prove useful in the analysis of other non-elitist algorithms as well; this has already been demonstrated very recently for a modified \ocl on the multimodal \textsc{Cliff} problem~\cite{Hevia2021FOGA}.

In Section~\ref{sec:experiments} we complement our runtime analyses with experiments on \onemax. First we compare the runtime of the \saocl, the self-adjusting \opl and the \ocl with the best known fixed~$\lambda$ for different problem sizes. Second, we show a sharp threshold for the success rate at $s\approx3.4$ where the runtime changes from polynomial to exponential. This indicates that the widely used one-fifth rule $(s = 4)$ is inefficient here, but other success rules achieve optimal asymptotic runtime. Finally, we show how different values of $s$ affect fixed target running times, the growth of~$\lambda$ over time and the time spent in each fitness value, shedding light on the non-optimal equilibrium states in the \saocl.

An extended abstract containing preliminary versions of our results appeared in \cite{Hevia2021}. The results in this manuscript have evolved significantly from there.
In~\cite{Hevia2021} we bounded the expected number of evaluations by showing that, when the fitness distance to the optimum has decreased below $n/\log^3 n$, the \saocl behaves similarly to its elitist version, a \saopl. The expected number of evaluations to reach this fitness distance was estimated using Wald's equation, and a reviewer of this manuscript pointed out a mistake in the application of Wald's equation in~\cite{Hevia2021} as the assumption of independent random variables was not met. We found a different argument to fix the proof and noticed that the new argument simplifies the analysis considerably. In particular, the simplified proof is no longer based on the elitist \saopl. (We remark that, independently, the analysis from~\cite{Hevia2021} was also simplified in~\cite{KaufmannArxiv2022,KaufmannPPSN2022} and extended to the class of monotone functions.) 
%Since we believe that the analysis of the \saopl is of independent interest, we analyse the \saopl in Section~\ref{sec:elitist-saopl}. Section~\ref{sec:evaluations} shows our simplified analysis that uses different arguments from the analysis of the elitist \saopl.

Other changes include rewriting our results in order to refine the presentation and give a more unified analysis for both our positive and negative results. 
In particular, the conditions on $s$ have been relaxed from $s < \frac{e-1}{e}$ vs.\ $s \ge 22$ towards $s < 1$ vs.\ $s \ge 18$.
We also extended our negative results towards other fitness function classes \jumpk, \cliffd, \zeromax, \twomax and \ridge (Theorem~\ref{thm:extended_functions}).

\section{Preliminaries}
\label{sec:Preliminaries}
We study the expected number of generations and fitness evaluations of the self-adjusting \ocl with self-adjusted offspring population size~$\lambda$ to find the optimum of the $n$-dimensional pseudo-Boolean function
${\onemax(x) := \sum_{i=1}^{n} x^{(i)}}$.
We define $X_0, X_1,\dots$ as the sequence of states of the algorithm, where $X_t = (x_t, \lambda_t)$ describes the current search point $x_t$ and the offspring population size $\lambda_t$ at generation~$t$. We often omit the subscripts $t$ when the context is obvious.

Using the naming convention from \cite{DoerrSurvey2020} we call the algorithm self-adjusting $(1,\{F^{1/s}\lambda, \lambda/F \})$~EA (Algorithm \ref{alg:saocl}). The algorithm behaves as the conventional \ocl: each generation it creates $\lambda$ offspring by flipping each bit independently with probability $1/n$ from the parent and selecting the fittest offspring as a parent for the next generation. In addition, in every generation it adjusts the offspring population size depending on the success of the generation. If the fittest offspring $y$ is better than the parent~$x$, the offspring population size is divided by the \emph{update strength} $F>1$, and multiplied by $F^{1/s}$ otherwise, with $s>0$ being the \emph{success rate}.

The idea of the parameter control mechanism is based on the interpretation of the one-fifth success rule from~\cite{KMH04}. The parameter $\lambda$ remains constant if the algorithm has a success every $s+1$ generations as then its new value is $\lambda \cdot (F^{1/s})^s \cdot 1/F = \lambda$. In pseudo-Boolean optimisation, the one-fifth success rule was first implemented by Doerr et al.~\cite{Doerr2015}, and proved to track the optimal offspring population size on the \ga in \cite{Doerr2018}. Our implementation is closer to the one used in \cite{Doerr2021}, where the authors generalise the success rule, implementing the success rate $s$ as a hyper-parameter.

Note that we regard $\lambda$ to be a real value, so that changes by factors of $1/F$ or $F^{1/s}$ happen on a continuous scale. Following Doerr and Doerr~\cite{Doerr2018}, we assume that, whenever an integer value of $\lambda$ is required, $\lambda$ is rounded to a nearest integer. For the sake of readability, we often write $\lambda$ as a real value even when an integer is required. Where appropriate, we use the notation $\round{\lambda}$ to denote the integer nearest to $\lambda$ (that is, rounding up if the fractional value is at least $0.5$ and rounding down otherwise).

\begin{algorithm}
\caption{Self-adjusting $(1,\{F^{1/s}\lambda, \lambda/F \})$~EA.}
\SetKwInput{Init}{Initialization}
\SetKwInput{Mut}{Mutation}
\SetKwInput{Sel}{Selection}
\SetKwInput{Opt}{Optimization}
\SetKwInput{Upd}{Update}
\label{alg:saocl}
\Init{Choose $x \in \{0,1\}^n$ uniformly at random (u.a.r.) and $\lambda := 1$\;}
\Opt{
	\For{$t \in \{1,2,\dots\}$}
		{
		\Mut{
			\For{$i \in \{1,\dots,\lambda \}$}
				{
				Create $y_i' \in \{0, 1\}^n$ by copying~$x$ and flipping each bit independently with probability $1/n$.\;
				}}
		\Sel{
			 Choose $y \in \{y_1', \dots , y_{\lambda}'\}$ with $f(y) = \max\{f(y_1'), \dots, f(y_\lambda')\}$ u.a.r.\;}
		\Upd{}
			\lIf{$f(y) > f(x)$}{ $x\leftarrow y$; $\lambda \leftarrow \max\{1, \lambda/F\}$}
			\lElse{
				$x\leftarrow y$; $\lambda \leftarrow F^{1/s}\lambda$}
		}
}
\end{algorithm}

\subsection{Notation and Probability Estimates}

We now give notation and tools for all \ocl algorithms.
\begin{definition}\label{def:probs-and-drifts}
For all $\lambda \in \mathbb{N}$ and $0 \le i < n$ we define:
\begin{align*}
    &\ploss = \prob{f(x_{t+1})<i\mid f(x_t)=i} &\\
    &\peq = \prob{f(x_{t+1})=i\mid f(x_t)=i} &\\
    &\pimp = \prob{f(x_{t+1})>i\mid f(x_t)=i} &\\
    &\Deltaloss = \E{i-f(x_{t+1})\mid f(x_t)=i \text{ and } f(x_{t+1})<i } &\\
    &\Deltagain = \E{f(x_{t+1})-i\mid f(x_t)=i \text{ and } f(x_{t+1})>i } &
\end{align*}
\end{definition}

As in~\cite{Rowe2014}, we call $\Deltagain$ \emph{forward drift} and $\Deltaloss$ \emph{backward drift} and note that they are both at least~1 by definition. We call the event underlying the probability $\ploss$ a \emph{fallback}, that is, the event that all offspring have lower fitness than the parent and thus $f(x_{t+1})< f(x_{t})$. The probability of a fallback, is $\ploss = (p_{i, 1}^-)^\lambda$ since all offspring must have worse fitness than their parent. Now, $p_{i, 1}^+$ is the probability of one offspring finding a better fitness value and $\pimp = {1 - (1-p_{i, 1}^+)^\lambda}$ since it is sufficient that one offspring improves the fitness.  Along with common bounds %$\frac{n-i}{en} \le p_{i, 1}^+ \le \frac{n-i}{n}$
and standard arguments, we obtain the following lemma.

\begin{lemma} \label{lem:bounds_probabilites}
For any \ocl on \onemax, the quantities from Definition~\ref{def:probs-and-drifts} are bounded as follows.
\begin{align}
    1-\frac{en}{en+\lambda(n-i)}
    \le 1-\left(1-\frac{n-i}{en}\right)^{\lambda} \!\! &\le\; \pimp
    \label{eq:pimp} \\
    \pimp \le 1-\left(1-1.14\left(\frac{n-i}{n}\right)\left(1-\frac{1}{n}\right)^{n-1}\right)^{\lambda}
     &\le 1-\left(1-\frac{n-i}{n} \right)^{\lambda} \label{eq:pimp2}
\end{align}
If $0.84n\le i\le 0.85n$ and $n\ge163$, then $p_{i,1}^+ \le 0.069$.
\begin{align}
    \left(\frac{i}{n}-\frac{1}{e}\right)^\lambda\le\; \ploss \le \left(1-\frac{n-i}{en}-\left(1-\frac{1}{n}\right)^n\right)^{\lambda}
    &\le \left(\frac{e-1}{e}\right)^{\lambda} \label{eq:ploss}
\end{align}
\vspace*{-\bigskipamount}
\begin{align}
    1 \le \Deltaloss \le\;& \frac{e}{e-1}\label{eq:expectedLoss}\\
    1 \le \Deltagain \le\;& \sum_{j=1}^\infty \left(1-\left(1-\frac{1}{j!}\right)^\lambda\right)\label{eq:expectedGain}
\end{align}
If $\lambda\ge 5$, then $\Deltagain \le \lceil\log\lambda\rceil + 0.413$.
\end{lemma}

\begin{proof}
We start by bounding the probability of one offspring being better than the parent $x$. For the lower bound a sufficient condition for the offspring to be better than the parent is that only one 0-bit is flipped. Therefore,
\begin{align}\label{eq:lower-bound-success-one-offspring}
    p_{i, 1}^+ \ge \frac{n-i}{n} \left(1-\frac{1}{n}\right)^{n-1} \ge \frac{n-i}{en}.
\end{align}

Along with $\pimp = 1-(1-p_{i,1}^+)^\lambda$, this proves one of the lower bounds in Equation~\eqref{eq:pimp} in Lemma \ref{lem:bounds_probabilites}.
Additionally, using $(1+x)^r \le \frac{1}{1 - rx}$ for all $x \in [-1,0]$ and $r \in \mathbb{N}$
\begin{align*}
    \pimp &\ge 1-\left(1-\frac{n-i}{en}\right)^{\lambda}
    \ge 1-\frac{1}{1+\frac{\lambda(n-i)}{en}}
    = 1-\frac{en}{en+\lambda(n-i)}.
\end{align*}

For the upper bound a necessary condition for the offspring to be better than the parent is that at least one 0-bit is flipped, hence
\begin{align*}
    p_{i, 1}^+ \le \frac{n-i}{n}.
\end{align*}

Additionally, we use the following upper bound shown in \cite{Paixao}:
\begin{align*}
    p_{i, 1}^+ \le \min \left\{1.14\left(\frac{n-i}{n}\right)\left(1-\frac{1}{n}\right)^{n-1}, 1\right\}.
\end{align*}
Since $1.14\left(\frac{n-i}{n}\right)\left(1-\frac{1}{n}\right)^{n-1}\le 1$ for all problem sizes $n>1$ and $n=1$ is trivially solved we omit the minimum from now on.

Along with $\pimp = 1-(1-p_{i,1}^+)^\lambda$, this proves the upper bounds in Equation~\eqref{eq:pimp2} in Lemma \ref{lem:bounds_probabilites}.

The additional upper bound for $p_{i,1}^+$ when $0.84n\le i\le0.85n$ uses a more precise bound from~\cite{Hevia2021FOGA} of:
\begin{align*}
    p_{i,1}^+ \le\;& \left(1- \frac{1}{n}\right)^{n-2}\sum_{a=0}^\infty \sum_{b=a+1}^\infty \left(\frac{i}{n}\right)^a \left(\frac{n-i}{n}\right)^b \frac{1}{a!b!}\\
    \le\;&\frac{1}{e}\left(1- \frac{1}{n}\right)^{-2} \sum_{a=0}^\infty \sum_{b=a+1}^\infty \left(0.85\right)^a \left(0.16\right)^b \frac{1}{a!b!}\\
    \le\;& \left(1- \frac{1}{n}\right)^{-2}
    0.068152.
\end{align*}
This implies that, for every $n\ge163$ and $0.84n\le i\le 0.85n$, $p_{i,1}^+\le 0.069$.

We now calculate $p_{i, 1}^-$. For the upper bound we use
\begin{align*}
    p_{i,1}^- = 1-p_{i, 1}^+-p_{i, 1}^0 .
\end{align*}
Using Equation~\eqref{eq:lower-bound-success-one-offspring} and bounding $p_{i, 1}^0$ from below by the probability of no bit flipping, that is,
\begin{align*}
    p_{i, 1}^0 \ge \left(1-\frac{1}{n}\right)^n,
\end{align*}
we get
\begin{align}\label{eq:upper-bound-loss-one-offspring}
    p_{i,1}^- \le 1- \frac{n-i}{en}- \left(1-\frac{1}{n}\right)^n.
\end{align}

Finally, for the lower bound we note that for an offspring to have less fitness than the parent it is sufficient that one of the $i$ 1-bits and none of the 0-bits is flipped. Therefore,
\begin{align}\label{eq:lower-bound-loss-one-offspring}
    p_{i,1}^-
    &\ge \left(1-\left(1-\frac{1}{n}\right)^{i}\right)\left(1-\frac{1}{n}\right)^{n-i}\nonumber\\
    &=\left(1-\frac{1}{n}\right)^{n-i} - \left(1-\frac{1}{n}\right)^{n}\nonumber\\
    &\ge1-\frac{n-i}{n}- \frac{1}{e}
    = \frac{i}{n}- \frac{1}{e}.
\end{align}
Using $\ploss = (p_{i,1}^-)^\lambda$ with Equations \eqref{eq:upper-bound-loss-one-offspring} and \eqref{eq:lower-bound-loss-one-offspring} we obtain
\begin{align*}
    \left(\frac{i}{n}-\frac{1}{e}\right)^\lambda\le\; \ploss \le \left(1- \frac{n-i}{en}-\left(1-\frac{1}{n}\right)^n\right)^{\lambda}.
\end{align*}
The upper bound is simplified as follows:
\begin{align*}
\ploss \le\;& \left(1- \frac{n-i}{en}-\left(1-\frac{1}{n}\right)^n\right)^{\lambda}\\
\le\;& \left(1- \frac{1}{en}-\left(1-\frac{1}{n}\right)^n\right)^{\lambda}\\
\le\;& \left(1- \frac{1}{en}-\frac{1}{e}\left(1-\frac{1}{n}\right)\right)^{\lambda}\\
=\;& \left(1-\frac{1}{e}\right)^{\lambda} = \left(\frac{e-1}{e}\right)^{\lambda}.
\end{align*}

To prove the bounds on the backward drift from Equation~\eqref{eq:expectedLoss}, note that the drift is conditional on a decrease in fitness, hence the lower bound of~1 is trivial.

The backward drift of a generation with $\lambda$ offspring can be upper bounded by a generation with only one offspring.

We pessimistically bound the backward drift by the expected number of flipping bits in a standard bit mutation. Under this pessimistic assumption, the condition $f(x_{t+1}) < i$ is equivalent to at least one bit flipping. Let $B$ denote the random number of flipping bits in a standard bit mutation with mutation probability $1/n$, then $\E{B} = 1$, $\Prob{B \ge 1} = 1-(1-1/n)^n \ge 1-1/e=(e-1)/e$ and
\begin{align*}
    \Deltaloss \le \E{B \mid B \ge 1} =\;& \sum_{x=1}^\infty \Prob{B = x \mid B \ge 1} \cdot x\\
    =\;& \sum_{x=1}^\infty \frac{\Prob{B = x}}{\Prob{B \ge 1}} \cdot x = \frac{\E{B}}{\Prob{B \ge 1}} \le
    \frac{e}{e-1}.
\end{align*}

The lower bound on the forward drift, Equation~\eqref{eq:expectedGain}, is again trivial since the forward drift is conditional on an increase in fitness.

To find the upper bound of $\Deltagain$ we pessimistically assume that all bit flips improve the fitness. Then we use the expected number of bit flips to bound $\Deltagain$. Let $B$ again denote the random number of flipping bits in a standard bit mutation with mutation probability $1/n$, then
\begin{align}
\label{eq:distribution-of-flipping-bits}
    \Prob{B\ge j}=\binom{n}{j}\left(\frac{1}{n}\right)^j\le \frac{1}{j!}.
\end{align}
To bound $\Deltagain$ we use the probability that any of the $\lambda$ offspring flip at least $j$ bits. Let $M_\lambda$ denote the maximum of the number of bits flipped in $\lambda$ independent standard bit mutations, then we have $\Prob{M_\lambda \ge j} = 1 - (1-\Prob{B \ge j})^\lambda$ and
\begin{align*}
    \Deltagain \le \E{M_\lambda} \le \sum_{j=1}^\infty \Prob{M_\lambda \ge j} \le \sum_{j=1}^\infty \left(1-\left(1-\frac{1}{j!}\right)^\lambda\right).
\end{align*}
For $\lambda\ge5$ we bound the first  $\lceil\log\lambda\rceil$ summands by 1 and apply Bernoulli's inequality:
\begin{align*}
    \Deltagain &\le \lceil\log\lambda\rceil +
    \sum_{j=\lceil\log\lambda\rceil+1}^\infty \left(1-\left(1-\frac{1}{j!}\right)^\lambda\right) \\
    &\le \lceil\log\lambda\rceil +
    \lambda \sum_{j=\lceil\log\lambda\rceil+1}^\infty \frac{1}{j!}\\
    &\le \lceil\log\lambda\rceil +
    2^{\lceil \log \lambda \rceil} \sum_{j=\lceil\log\lambda\rceil+1}^\infty \frac{1}{j!}.
\end{align*}
The function $f\colon \mathbb{N} \to \mathbb{R}$ with $f(x) \coloneqq 2^x \sum_{j=x+1}^\infty \frac{1}{j!}$ is decreasing with~$x$ and thus for all $\lambda \ge 5$ we get $\Deltagain \le \lceil \log \lambda \rceil + f(3) = \lceil \log \lambda \rceil + \frac{8}{3}(3e-8) < \lceil \log \lambda \rceil + 0.413$.
\end{proof}

We now show the following lemma that establishes a natural limit to the value of $\lambda$.

\begin{lemma}\label{lem:probability-exceed-polyn}
Consider the \saocl on any unimodal function with an initial offspring population size of $\lambda_0 \le eF^{1/s}n^3$. The probability that, during a run, the offspring population size exceeds ${e F^{1/s} n^3}$ before the optimum is found is at most $\exp(-\Omega(n^2))$.
\end{lemma}

\begin{proof}
In order to have ${\lambda_{t+1} \ge e F^{1/s} n^3}$, a generation with ${\lambda_t \ge en^3}$ must be unsuccessful. Since there is always a one-bit flip that improves the fitness and the probability that an offspring flips only one bit is $\frac{1}{n} \left(1-\frac{1}{n}\right)^{n-1} \ge \frac{1}{en}$, then the probability of an unsuccessful generation
with $\lambda \ge en^3$ is at most
\begin{align*}
    \left(1-\frac{1}{en}\right)^{e n^3}
    \le \exp(-n^2).
\end{align*}

The probability of finding the optimum in one generation with any $\lambda$ and any current fitness is at least $n^{-n}=\exp(-n\ln n)$. Hence the probability of exceeding ${\lambda=e F^{1/s} n^3}$ before finding the optimum is at most
\begin{align*}
    \frac{\exp(-n^2)}{\exp(-n\ln n)+\exp(-n^2)}
    &\le\frac{\exp(-n^2)}{\exp(-n\ln n)}
    =\exp(-\Omega(n^2)).
\end{align*}
\end{proof}

\subsection{Drift Analysis and Potential Functions}

Drift analysis is one of the most useful tools to analyse evolutionary algorithms~\cite{Lengler2020}. A general approach for the use of drift analysis is to identify a potential function that adequately captures the progress of the algorithm and the distance from a desired target state (e.\,g.\ having found a global optimum). Then we analyse the expected changes in the potential function at every step of the optimisation (drift of the potential) and finally translate this knowledge about the drift into information about the runtime of the algorithm.

Several powerful drift theorems have been developed throughout the years that help with the last step of the above approach, requiring as little information as possible about the potential and its drift. Hence, this step is relatively straightforward. For convenience, we state the drift theorems used in our work.

\begin{theorem}[Additive Drift~\cite{He2004}]
	\label{thm:additive_drift}
	Let	$(X_t)_{t\geq 0}$ be a sequence of non-negative random variables over a finite state space $S \subseteq \mathbb{R}$. Let $T$ be the random variable that denotes the earliest point in time $t\geq 0$ such that $X_t = 0$. If there exists $c > 0$ such that, for all $t<T$,
	\[
		\E{X_{t} - X_{t+1} \mid X_t} \geq c ,
	\]
	then
	\[
		\E{T \mid X_0} \leq \frac{X_0}{c} .
	\]
\end{theorem}

The following two theorems both deal with the case that the drift is pointing away from the target, that is, the expected progress is negative in an interval of the state space.
\begin{theorem}[Negative drift theorem~\cite{OlivetoNegDrift,Oliveto2012Erratum}]
\label{thm:negative-drift}
Let $X_t$, $t \ge 0$, be real-valued random variables describing a stochastic process over some state space. Suppose there exists an interval $[a, b] \subseteq \mathbb{R}$, two constants $\delta, \varepsilon > 0$ and, possibly depending on $\ell \coloneqq b-a$, a function $r(\ell)$ satisfying $1 \le r(\ell) = o(\ell/\log(\ell))$ such that for all $t \ge 0$ the following two conditions hold:
\begin{enumerate}
    \item $\E{X_{t+1} - X_t \mid X_0, \dots, X_t; a < X_t < b} \ge \varepsilon$.
    \item $\Prob{\lvert X_{t+1} - X_t\rvert \ge j \mid X_0, \dots, X_t; a < X_t} \le \frac{r(\ell)}{(1+\delta)^j}$ for $j \in \mathbb{N}_0$.
\end{enumerate}
Then there exists a constant $c^* > 0$ such that for ${T^* \coloneqq \min\{t \ge 0 \colon X_t < a} \mid X_0, \dots, X_t; X_0 \ge b\}$ it holds $\Prob{T^* \le 2^{c^*\ell/r(\ell)}} = 2^{-\Omega(\ell/r(\ell))}$.
\end{theorem}

The following theorem is a variation of Theorem~\ref{thm:negative-drift} in which the second condition on large jumps is relaxed.
\begin{theorem}[Negative drift theorem with scaling~\cite{Oliveto2015}]
\label{thm:negative-drift-with-scaling}
Let $X_t$, $t > 0$ be real-valued random variables describing a stochastic process over some state space. Suppose there exists an interval $[a, b] \subseteq \mathbb{R}$ and, possibly depending on $\ell := b-a$, a drift bound $\varepsilon \coloneqq \varepsilon(\ell) > 0$ as well as a scaling factor $r \coloneqq r(\ell)$ such that for all $t \ge 0$ the following three conditions hold:
\begin{enumerate}
    \item $\E{X_{t+1} - X_t \mid X_0, \dots, X_t; a < X_t < b} \ge \varepsilon$.
    \item $\Prob{\lvert X_{t+1} - X_t\rvert \ge jr \mid X_0, \dots, X_t; a < X_t} \le e^{-j}$ for $j \in \mathbb{N}_0$.
    \item $1 \le r^2 \le \varepsilon \ell/(132\log(r/\varepsilon))$.
\end{enumerate}
Then for the first hitting time ${T^* \coloneqq \min\{t \ge 0 \colon X_t < a} \mid X_0, \dots, X_t; X_0 \ge b\}$ it holds that $\Prob{T^* \le e^{\varepsilon \ell/(132r^2)}} = O(e^{-\varepsilon \ell/(132r^2)})$.
\end{theorem}

For our analysis the first step, that is, finding a good potential function is much more interesting. A natural candidate for a potential function is the fitness of the current individual $f(x_t)$. However, the \saocl adjusts $\lambda$ throughout the optimisation, and the expected change in fitness crucially depends on the current value of~$\lambda$. Therefore, we also need to take into account the current offspring population size~$\lambda$ and capture both fitness and $\lambda$ in our potential function. Since we study different behaviours of the algorithm depending on the success rate $s$ we generalise the potential function used in~\cite{Hevia2021} by considering an abstract function $h(\lambda_t)$ of the current offspring population sizes. The function $h(\lambda_t)$ will be chosen differently for different contexts, such as proving a positive result for small success rates~$s$ and proving a negative result for large success rates.

\begin{definition}\label{def:potential-function}
Given a function $h \colon \mathbb{R} \to \mathbb{R}$, we define the potential function $g(X_t)$ as
\begin{align*}
    g(X_t) = f(x_t) + h(\lambda_t).
\end{align*}
\end{definition}
We do not make any assumptions on $h(\lambda_t)$ at this stage, but we will choose $h(\lambda_t)$ in the following sections as functions of $\lambda_t$ that reward increases of $\lambda_t$, for small values of~$\lambda_t$.
%describes the progress of the algorithm towards the desired or expected values of $\lambda_t$.
%
We note that this potential function is also a generalisation of the potential function used in~\cite{Hevia2021FOGA} to analyse the \saocl with a reset mechanism on the \cliff function. We believe that this approach could be useful for the analysis of a wide range of success-based parameter control mechanisms and it might be able to simplify previous analysis such as~\cite{Doerr2018,Doerr2021}. A similar approach has been used before for continuous domains in~\cite{Akimoto2018,Akimoto2019,Akimoto2021}.

For every function $h(\lambda_t)$, we can compute the drift in the potential as shown in the following lemma. For the sake of readability we drop the subscript $t$ in $\lambda_t$ where appropriate.

\begin{lemma}\label{lem:generalised-potential}
Consider the \saocl.
Then for every function $h \colon \mathbb{R} \to \mathbb{R}^+_0$ and every generation $t$ with $f(x_t) < n$ and $\lambda_t>F$, $\E{g(X_{t+1})-g(X_{t})\mid X_{t}}$ is
\[
    \left(\Deltagain+h(\lambda/F)-h(\lambda F^{1/s})\right)\pimp + h(\lambda F^{1/s}) -h(\lambda) -\Deltaloss \ploss.
\]
If $\lambda_t\le F$ then, $\E{g(X_{t+1})-g(X_{t})\mid X_{t}}$ is
\[
    \left(\Deltagain+h(1)-h(\lambda F^{1/s})\right)\pimp + h(\lambda F^{1/s}) -h(\lambda) -\Deltaloss \ploss.
\]
\end{lemma}
\begin{proof}
When an improvement is found, the fitness increases in expectation by $\Deltagain$ and since $\lambda_{t+1}=\lambda/F$, the $\lambda$~term changes by $h(\lambda/F)-h(\lambda)$. When the fitness does not change, the $\lambda$~term changes by $h(\lambda F^{1/s})-h(\lambda)$. When the fitness decreases the expected decrease is $\Deltaloss$ and the $\lambda$~term changes by $h(\lambda F^{1/s})-h(\lambda)$. Together $\E{g(X_{t+1})-g(X_{t})\mid X_{t}}$ is
\begin{align*}
    &\left(\Deltagain + h(\lambda/F) - h(\lambda)\right) \pimp + \left(h(\lambda F^{1/s})-h(\lambda)\right)\peq +\\
    &\hspace{6.5cm}\left(h(\lambda F^{1/s})-h(\lambda)-\Deltaloss\right)\ploss\\
    &=\left(\Deltagain + h(\lambda/F) - h(\lambda)\right) \pimp + \left(h(\lambda F^{1/s})-h(\lambda)\right)(\peq+\ploss) -\Deltaloss\ploss\\
    &=\left(\Deltagain + h(\lambda/F) - h(\lambda)\right) \pimp + \left(h(\lambda F^{1/s})-h(\lambda)\right)(1-\pimp) -\Deltaloss\ploss\\
    &=\left(\Deltagain + h(\lambda/F) - h(\lambda F^{1/s})\right) \pimp + h(\lambda F^{1/s})-h(\lambda) -\Deltaloss\ploss
\end{align*}
Given that $\lambda\ge1$ if $\lambda\le F$ then $h(\lambda/F)$ needs to be replaced by $h(\lambda/\lambda)=h(1)$.
\end{proof}

\section{Analysing the Elitist \saopl}
\label{sec:elitist-saopl}

We first consider the method of amortised analysis to analyse self-adjusting EAs. As explained in the introduction, this approach only applies to elitist algorithms. We therefore consider the elitist version of the \saocl, the \saopl, and bound its expected time to go from $a$ to~$b$ for arbitrary fitness thresholds $a < b$. We believe that this section and the analysis of the \saopl is of independent interest, even though our main results on the non-elitist \saocl use different arguments presented in Section~\ref{sec:evaluations}. The results from this section have already been used in follow-up work~\cite{Hevia2022} and the proof technique, a refinement and generalisation of the accounting method used in~\cite{Lassig2011}, may find further applications in the context of self-adjusting parameters. 

The following statement generalises results from~\cite{Lassig2011}, which considered hard-wired parameters $F=2, s=1$ to arbitrary update strengths~$F > 1$, arbitrary success rates $s > 0$, arbitrary initial values of the offspring population size and arbitrary fitness intervals $[a, b]$. The latter is directly applicable for fixed-target analyses~\cite{BuzdalovDDV22} where one is looking to determine the expected time until an algorithm achieves a given fitness threshold.

\begin{theorem}
\label{thm:runtime-of-elitist-algorithm}
Consider the elitist \saopl on \onemax with ${f(x_0) \ge a}$ and an initial offspring population size of~$\lambda_0$. For every integer~$b \le n$, the expected number of evaluations for it to reach a fitness of at least~$b$ is at most
% \[
%     \lambdainit \cdot \frac{F}{1-F} + n \cdot \frac{(2+2e)F^{1/s}-1}{F^{1/s} - 1} \cdot \frac{F^{\frac{s+1}{s}}-1}{F-1} \sum_{j=a}^{b - 1} \frac{1}{n-j}.
% \]
% \[
%     \lambdainit \cdot \frac{F}{1-F} + \left(\frac{1}{e} + \frac{1}{\ln(F^{1/s})} \cdot \left(1 - F^{-1/s}\right)\right) \cdot \frac{F^{\frac{s+1}{s}}-1}{F-1} \sum_{j=a}^{b - 1} \frac{1}{p_{i,1}^+}.
% \]
% \[
%     \lambdainit \cdot \frac{F}{1-F} + \left(1 + \frac{e}{\ln(F^{1/s})} \cdot \left(1 - F^{-1/s}\right)\right) \cdot \frac{F^{\frac{s+1}{s}}-1}{F-1} \sum_{j=a}^{b - 1} \frac{n}{n-j}.
% \]
\[
    \lambdainit \cdot \frac{F}{F-1} + \left(\frac{1}{e} + \frac{1 - F^{-1/s}}{\ln(F^{1/s})}\right) \cdot \frac{F^{\frac{s+1}{s}}-1}{F-1} \sum_{i=a}^{b - 1} \frac{en}{n-i}.
\]
\end{theorem}
Note that the term $\sum_{i=a}^{b - 1} \frac{en}{n-i}$ represents an upper bound for the expected time for the (1+1)~EA on \textsc{OneMax}, starting with a fitness of~$a$, to reach a fitness of at least~$b$ obtained via the fitness-level method (where the highest fitness level contains all search points with fitness at least~$b$). This bound is known to be tight up to small-order terms for $b=n$~\cite{Sudholt13,DoerrKoetzing2021}. The bound from Theorem~\ref{thm:runtime-of-elitist-algorithm} is thus only by a constant factor and an additive term of $\lambdainit \cdot \frac{F}{1-F}$ larger.

We further remark that Theorem~\ref{thm:runtime-of-elitist-algorithm}, applied with $a := 0$ and $b := n$, immediately implies the following.
\begin{corollary}
The expected number of function evaluations of the elitist \mbox{\saopl} using any constant parameters $s > 0$, $F > 1$ and ${\lambda_0 = O(n \log n)}$ on \onemax is $O(n \log n)$.
\end{corollary}

To prove Theorem~\ref{thm:runtime-of-elitist-algorithm}, we define a fitness level $i$ as the set of all search points $x$ with fitness $f(x)=i$ and argue that the analysis can be boiled down to the number of evaluations spent in each fitness level \emph{during unsuccessful generations}. This is made precise in the following lemma.
\begin{lemma}
\label{lem:abstract-bound-from-accounting-method}
Consider the elitist \saopl on \onemax with ${f(x_0) \ge a}$ and an initial offspring population size of~$\lambda_0$. Fix an integer~$b \le n$ and, for all $0 \le i \le n-1$, let $U_i$ denote the number of function evaluations made during all unsuccessful generations on fitness level~$i$.
Then the number of evaluations to reach a fitness of at least~$b$ is at most
\[
\lambdainit \cdot \frac{F}{F-1} + \frac{F^{\frac{s+1}{s}}-1}{F-1} \sum_{i=a}^{b - 1} U_i.
\]
A bound on the \emph{expected} number of evaluations is obtained by replacing $U_i$ with $\E{U_i}$.
\end{lemma}
\begin{proof}
We refine the \emph{accounting method} used in the analysis of the $(1+\{2\lambda, \lambda/2\})$~EA in~\cite{Rowe2014}. The main idea is: if some fitness level~$i$ increases $\lambda$ to a large value, we charge the costs for increasing $\lambda$ to that fitness level. In addition, we charge costs that pay for decreasing $\lambda$ down to~1 in future successful generations. Hence, a successful generation comes for free, at the expense of a constant factor for the cost of unsuccessful generations.

Let $\phi(\lambda_t) := \frac{F}{F-1} \lambda_t$ and imagine a fictional bank account. We make an initial payment of $\phi(\lambdainit)$ to that bank account. If a generation~$t$ is unsuccessful, we pay $\lambda_t$ for the current generation and deposit an additional amount of $\phi(\lambda_t F^{1/s}) - \phi(\lambda_t)$. If an improvement is found, we withdraw an amount of $\lambda_t$ to pay for generation~$t$.

We show by induction: for every generation~$t$, the account's balance is $\phi(\lambda_t)$.
This is true for the initial generation owing to the initial payment of $\phi(\lambdainit)$.
Assume the statement holds for time~$t$. If generation~$t$ is unsuccessful, the new balance is
\[
    \phi(\lambda_t) + (\phi(\lambda_t F^{1/s}) - \phi(\lambda_t)) = \phi(\lambda_t F^{1/s}) = \phi(\lambda_{t+1}).
\]
If generation~$t$ is successful, $\lambda_{t+1}=\lambda_t/F$ and  the new balance is
\[
    \phi(\lambda_t) - \lambda_t = \left(\frac{F}{F-1} - 1\right)\lambda_t = \frac{1}{F-1} \cdot \lambda_t = \frac{F}{F-1} \cdot \lambda_{t+1} = \phi(\lambda_{t+1}).
\]
Now the costs of an unsuccessful generation~$t$ are
\[
    \lambda_t + \phi(\lambda_t F^{1/s}) - \phi(\lambda_t)
    = \lambda_t \left(1 + \frac{F^{\frac{s+1}{s}}}{F-1} - \frac{F}{F-1}\right) = \lambda_t \cdot \frac{F^{\frac{s+1}{s}}-1}{F-1},
\]
that is, by a factor of $\frac{F^{\frac{s+1}{s}}-1}{F-1}$ larger than the number of evaluations in that generation. Recall that successful generations incur no costs as the additional factor in unsuccessful generation has already paid for these evaluations. This implies that, if $U_i$ denotes the number of evaluations during all unsuccessful generations on fitness level~$i$, the costs incurred on fitness level~$i$ are $U_i \cdot \frac{F^{\frac{s+1}{s}}-1}{F-1}$. Summing up over all non-optimal~$i$ and adding costs $\phi(\lambdainit) = \lambdainit \cdot \frac{F}{F-1}$ to account for the initial value of~$\lambda$ yields the claimed bound.

The final statement on the \emph{expected} number of evaluations follows from taking expectations and exploiting linearity of expectations.
\end{proof}

The next step is to bound $\E{U_i}$, that is, the expected number of evaluations \emph{in unsuccessful generations} on an arbitrary but fixed fitness level~$i$.
\begin{lemma}
\label{lem:evaluations-in-unsuccessful-generations}
Consider the \saopl starting on fitness level~$i$ with an offspring population size of~$\lambda$.
For every initial~$\lambda$, the expected number of evaluations in unsuccessful generations for the \saopl on fitness level~$i$ is at most
% \[
%   \frac{1}{ep_{i,1}^+} +  \frac{s}{p_{i,1}^+ \ln(F)} \cdot \left(1 - F^{-1/s}\right).
% \]
\[
   \E{U_i} \le \frac{1}{p_{i,1}^+} \cdot \left(\frac{1}{e} + \frac{1-F^{-1/s}}{\ln(F^{1/s})}\right).
\]
\end{lemma}
Note that the bound from Lemma~\ref{lem:evaluations-in-unsuccessful-generations} does not depend on the initial value of~$\lambda$. Roughly speaking, for small values of~$\lambda$, the algorithm will typically increase $\lambda$ to a value where improvements become likely, and then the number of evaluations essentially depends on the difficulty of the fitness level, expressed through the factor of $\frac{1}{p_{i,1}^+}$ from the lemma's statement. If $\lambda$ is larger than required, with high probability the first generation is successful and then there are no evaluations in an unsuccessful generation on fitness level~$i$.

In order to prove Lemma~\ref{lem:evaluations-in-unsuccessful-generations}, we first need two technical lemmas. Their proofs can be found in the appendix. The first lemma bounds a sum by an integral, plus the largest possible function value.

\begin{theoremEnd}[restate, no link to proof]{lemma}
\label{lem:bounding-sum-by-interval-concave-function}
Let $f \colon \mathbb{R}_0^+ \to \mathbb{R}_0^+$ be an integrable function with a unique maximum at~$\alpha$. Then
\[
    \sum_{i=0}^\infty f(i) \le f(\alpha) + \int_{0}^\infty f(i) \; \mathrm{d}i.
\]
\end{theoremEnd}
\begin{proofEnd}
We assume that $\int_{0}^\infty f(i) \; \mathrm{d}i < \infty$ as otherwise the claim is trivial.
Since $f$ is non-decreasing in $[0, \lfloor \alpha \rfloor]$, for all $i = 0, \dots, {\lfloor \alpha \rfloor -1}$ we have $f(i) \le \int_{i}^{i+1} f(i) \; \mathrm{d}i$. This yields
\[
    \sum_{i=0}^{\lfloor \alpha \rfloor} f(i) \le \int_{0}^{\lfloor \alpha \rfloor} f(i) \; \mathrm{d}i + f(\lfloor \alpha \rfloor).
\]
Likewise, since $f$ is non-increasing in $[\lfloor \alpha \rfloor +1, \infty)$, for all $i = \lfloor \alpha \rfloor +2, \dots$ we have $f(i) \le \int_{i-1}^i f(i) \; \mathrm{d}i$. This yields
\[
    \sum_{i=\lfloor \alpha \rfloor +1}^{\infty} f(i) \le \int_{\lfloor \alpha \rfloor +1}^{\infty} f(i) \; \mathrm{d}i + f(\lfloor \alpha \rfloor).
\]
Assume $f(\lfloor \alpha \rfloor) \le f(\lfloor \alpha \rfloor + 1)$, then $f(i) \ge f(\lfloor \alpha \rfloor)$ for all $i \in [\lfloor \alpha \rfloor, \lfloor \alpha \rfloor + 1]$ and thus $f(\lfloor \alpha \rfloor) \le \int_{\lfloor \alpha \rfloor}^{\lfloor \alpha \rfloor + 1} f(i) \; \mathrm{d}i$.
This implies
\[
    f(\lfloor \alpha \rfloor) + f(\lfloor \alpha \rfloor + 1) \le f(\alpha) + \int_{\lfloor \alpha \rfloor}^{\lfloor \alpha \rfloor + 1} f(i) \; \mathrm{d}i.
\]
The case $f(\lfloor \alpha \rfloor) > f(\lfloor \alpha \rfloor + 1)$ is symmetric and leads to the same statement.
Together, this implies the claim.
\end{proofEnd}

We will also need a closed form for the following integral.
\begin{theoremEnd}[restate, no link to proof]{lemma}
\label{lem:closed-form-for-integral}
\[
    \int_{j=0}^\infty \alpha^{j} \cdot \exp\left(- \beta \cdot \alpha^j\right) \; \mathrm{d}j
    = \frac{e^{-\beta}}{\beta \ln(\alpha)}.
\]
\end{theoremEnd}
\begin{proofEnd}
The integral can be written as
\[
    \frac{1}{\ln(\alpha)} \int_{j=0}^\infty \alpha^{j} \ln(\alpha) \cdot \exp\left(- \beta \cdot \alpha^j\right) \; \mathrm{d}j.
\]
For $f(x) := \exp(-\beta x)$ and $\varphi(x) = \alpha^x$, along with $\varphi'(x) = \alpha^x \ln(\alpha)$, this is
\[
    \frac{1}{\ln(\alpha)} \int_{j=0}^\infty \varphi'(x) \cdot f(\varphi(x)) \; \mathrm{d}x.
\]
The rule of integration by substitution states that $\int_{a}^b \varphi'(x) \cdot f(\varphi(x)) \; \mathrm{d}x = \int_{\varphi(a)}^{\varphi(b)} f(u) \; \mathrm{d}u$. Using $\int \exp(-\beta u) \;\mathrm{d}u = -\frac{e^{-\beta u}}{\beta}+C$, the above is
\begin{align*}
    \frac{1}{\ln(\alpha)} \cdot \lim_{b \to \infty} \int_{\varphi(0)}^{\varphi(b)} f(u) \; \mathrm{d}u
    =\;& \frac{1}{\ln(\alpha)} \cdot \lim_{u \to \infty} \frac{e^{-\beta}-e^{-\beta u}}{\beta}
    = \frac{1}{\ln(\alpha)} \cdot  \frac{e^{-\beta}}{\beta}.
\end{align*}
\end{proofEnd}

Now we use these technical lemmas to prove Lemma~\ref{lem:evaluations-in-unsuccessful-generations}.
\begin{proof}[Proof of Lemma~\ref{lem:evaluations-in-unsuccessful-generations}]
Since we are considering an elitist algorithm, fitness level~$i$ is left for good once we have a success from a current search point on level~$i$. Starting with an offspring population size of~$\lambda$,
if generations $0, \dots, j$ are unsuccessful, the $j$-th unsuccessful generation has an offspring population size of $\lambda F^{j/s}$. However, it is only counted in the expectation we are aiming to bound if there is no success in generations $0, \dots, j$. There is no success in generations $0, \dots, j$ if and only if the algorithm makes $\lambda \sum_{\ell=0}^j F^{\ell/s} = \lambda \frac{F^{\frac{j+1}{s}}-1}{F^{1/s}-1}$ evaluations without generating an improvement. Hence the expected number of evaluations in unsuccessful generations is equal to
\begin{align*}
    & \sum_{j=0}^\infty \lambda F^{j/s} \cdot \Prob{\text{no success in $\lambda \cdot \frac{F^{\frac{j+1}{s}}-1}{F^{1/s}-1}$ evaluations}}\\
    &=\; \sum_{j=0}^\infty \lambda F^{j/s} \cdot \left(1 - p_{i,1}^+\right)^{\lambda \cdot \frac{F^{\frac{j+1}{s}}-1}{F^{1/s}-1}}\\
    &\le\; \sum_{j=0}^\infty \lambda F^{j/s} \cdot \exp\left(- p_{i,1}^+\lambda \cdot \frac{F^{\frac{j+1}{s}}-1}{F^{1/s}-1}\right).
\end{align*}
We will use Lemma~\ref{lem:bounding-sum-by-interval-concave-function} to bound the above sum by an integral and the maximum of the function $\lambda F^{j/s} \cdot \exp\left(- p_{i,1}^+\lambda \cdot \frac{F^{\frac{j+1}{s}}-1}{F^{1/s}-1}\right)$.

We define the (simpler) function $\xi(x) := x \cdot \exp(-p_{i,1}^+ x)$ and note that its maximum value is $1/(ep_{i,1}^+)$. This value is an upper bound for the sought maximum since
 $\frac{F^{\frac{j+1}{s}}-1}{F^{1/s}-1}=\sum_{\ell=0}^j F^{\ell/s} \ge F^{j/s}$ and thus
%  $\sum_{\ell=0}^j F^{\ell/s} = \frac{F^{\frac{j+1}{s}}-1}{F^{1/s}-1}$ implies $\frac{F^{\frac{j+1}{s}}-1}{F^{1/s}-1} \ge F^{j/s}$ and thus
\[
    \lambda F^{j/s} \cdot \exp\left(- p_{i,1}^+\lambda \cdot \frac{F^{\frac{j+1}{s}}-1}{F^{1/s}-1}\right)
    \le \lambda F^{j/s} \cdot \exp\left(-p_{i,1}^+\lambda F^{j/s}\right) = \xi(\lambda F^{j/s}) \le \frac{1}{ep_{i,1}^+}.
\]
%
% Hence the sought maximum is also bounded by $1/$
% Since
% \[
%     \xi(\lambda F^{j/s}) = \lambda F^{j/s} \cdot \exp\left(-p_{i,1}^+\lambda F^{j/s}\right) \ge \lambda F^{j/s} \cdot \exp\left(- p_{i,1}^+\lambda \cdot \frac{F^{\frac{j+1}{s}}-1}{F^{1/s}-1}\right)
% \]
% the value $1/(ep_{i,1}^+)$ is an upper bound for the maximum of
% $\exp\left(- p_{i,1}^+\lambda \cdot \frac{F^{\frac{j+1}{s}}-1}{F^{1/s}-1}\right)$.
Plugging this into the bound obtained by invoking Lemma~\ref{lem:bounding-sum-by-interval-concave-function}, the sum is thus at most
\begin{align}
    & \frac{1}{ep_{i,1}^+} + \int_{j=0}^\infty \lambda F^{j/s} \cdot \exp\left(- p_{i,1}^+\lambda \cdot \frac{F^{\frac{j+1}{s}}-1}{F^{1/s}-1}\right) \; \mathrm{d}j\notag\\
    &=\; \frac{1}{ep_{i,1}^+} + \exp\left(\frac{p_{i,1}^+\lambda}{F^{1/s}-1}\right) \int_{j=0}^\infty \lambda F^{j/s} \cdot \exp\left(- p_{i,1}^+\lambda \cdot \frac{F^{\frac{j+1}{s}}}{F^{1/s}-1}\right) \; \mathrm{d}j\label{eq:nonsimplified-formula}
\end{align}

Let $\alpha := F^{1/s}$ and $\beta := p_{i,1}^+ \lambda \cdot \frac{F^{1/s}}{F^{1/s}-1}$, then we can write
\[
    \int_{j=0}^\infty \lambda F^{j/s} \cdot \exp\left(- p_{i,1}^+\lambda \cdot \frac{F^{\frac{j+1}{s}}}{F^{1/s}-1}\right) \; \mathrm{d}j
    = \lambda \int_{j=0}^\infty \alpha^j \cdot \exp(-\beta \cdot \alpha^j).
\]
By Lemma~\ref{lem:closed-form-for-integral}, this equals
\[
    \lambda \cdot \frac{e^{-\beta}}{\beta \ln(\alpha)} = \frac{\exp\left(-p_{i,1}^+\lambda \cdot \frac{F^{1/s}}{F^{1/s}-1}\right)}{p_{i,1}^+ \cdot \frac{F^{1/s}}{F^{1/s}-1} \ln\left(F^{1/s}\right)}
    =
    \frac{\exp\left(-p_{i,1}^+\lambda \cdot \frac{F^{1/s}}{F^{1/s}-1}\right)(F^{1/s}-1)}{p_{i,1}^+ \cdot F^{1/s} \ln\left(F^{1/s}\right)}.
\]
Plugging this back into~\eqref{eq:nonsimplified-formula} yields
\begin{align*}
    & \frac{1}{ep_{i,1}^+} + \exp\left(\frac{p_{i,1}^+\lambda}{F^{1/s}-1}\right) \cdot
    \frac{\exp\left(-p_{i,1}^+\lambda \cdot \frac{F^{1/s}}{F^{1/s}-1}\right)(F^{1/s}-1)}{p_{i,1}^+ \cdot F^{1/s} \ln\left(F^{1/s}\right)}\\
    &=\; \frac{1}{ep_{i,1}^+} + \exp\left(\frac{p_{i,1}^+\lambda}{F^{1/s}-1}\cdot (1-F^{1/s})\right) \cdot
    \frac{F^{1/s}-1}{p_{i,1}^+ \cdot F^{1/s} \ln\left(F^{1/s}\right)}\\
    &=\;     \frac{1}{ep_{i,1}^+} + \exp\left(-p_{i,1}^+\lambda\right) \cdot
    \frac{F^{1/s}-1}{p_{i,1}^+ \cdot F^{1/s} \ln\left(F^{1/s}\right)}.
\end{align*}
In this upper bound, we can see that the worst value for $\lambda$ is $\lambda=1$.
Using $p_{i,1}^+\lambda \ge 0$ for all~$\lambda$ and thus bounding $\exp(-p_{i, 1}^+\lambda) \le 1$, we get an upper bound of
\begin{align*}
     \frac{1}{ep_{i,1}^+} + \frac{F^{1/s}-1}{p_{i,1}^+ F^{1/s} \ln(F^{1/s})}
     =  \frac{1}{ep_{i,1}^+} + \frac{1-F^{-1/s}}{p_{i,1}^+ \ln(F^{1/s})}
     =  \frac{1}{p_{i,1}^+} \cdot \left(\frac{1}{e} + \frac{1-F^{-1/s}}{\ln(F^{1/s})}\right).
\end{align*}
\end{proof}

Now proving Theorem~\ref{thm:runtime-of-elitist-algorithm} is quite straightforward.
\begin{proof}[Proof of Theorem~\ref{thm:runtime-of-elitist-algorithm}]
Combining Lemma~\ref{lem:abstract-bound-from-accounting-method} with Lemma~\ref{lem:evaluations-in-unsuccessful-generations} yields an upper bound of
\[
    \lambdainit \cdot \frac{F}{1-F} + \left(\frac{1}{e} + \frac{1-F^{-1/s}}{\ln(F^{1/s})}\right) \cdot \frac{F^{\frac{s+1}{s}}-1}{F-1} \sum_{i=a}^{b - 1} \frac{1}{p_{i,1}^+}.
\]
Plugging in $p_{i, 1}^+ \ge (n-i)/(en)$ yields the claimed bound.
\end{proof}

\section{Small Success Rates are Efficient}\label{sec:poly-runtime}

Now we consider the non-elitist \saocl and show that, for suitable choices of the success rate~$s$ and constant update strength~$F$, the \saocl optimises \onemax in  $O(n)$ expected generations and $O(n\log n)$ expected evaluations.
This section uses different arguments from those used in Section~\ref{sec:elitist-saopl} to analyse the \saocl's elitist counterpart, the \saopl.

\subsection{Bounding the Number of Generations}
\label{sec:generations}

We first only focus on the expected number of generations as the number of function evaluations depends on the dynamics of the offspring population size over time and is considerably harder to analyse. The following theorem states the main result of this section.
\begin{theorem}\label{thm:optimisationTime-generations}
Let the update strength $F>1$ and the success rate $0<s<1$ be constants. Then for any initial search point and any initial $\lambda$ the expected number of generations of the \saocl on \onemax is $O(n)$.
\end{theorem}

We note that the self-adjusting mechanism aims to obtain one success every $s+1$ generations. The intuition behind using $0<s<1$ in Theorem~\ref{sec:generations} is that then the algorithm tries to succeed (improve the fitness) more than half of the generations. In order to achieve that many successes the $\lambda$-value needs to be large, which in turn reduces the probability (and number) of fallbacks during the run.

% We note that the self-adjusting mechanism aims to obtain one success every $s+1$ generations. Theorem~\ref{sec:generations} uses $0<s<1$; for these parameter values the algorithm tries to succeed more than half of the generations. The intuition behind using $s<1$ is that in order to have that many successes the self-adjusting mechanism needs large $\lambda$-values, which in turn reduces the probability (and number) of fallbacks during the run.

% We note that Theorem~\ref{sec:generations} uses $0<s<1$. The intuition behind using $s<1$ is that the self-adjusting mechanism tries to maintain a $\lambda$-value large enough, such that more than half of the generations are successful. More importantly, these large $\lambda$-values also reduce the number of fallbacks.

We make use of the potential function from Definition~\ref{def:potential-function} and define $h(\lambda)$ to obtain the potential function used in this section as follows.

\begin{definition}\label{def:potential-function-pos}
We define the potential function $g_1(X_t)$ as
\begin{align*}
    g_1(X_t) = f(x_t) - \frac{2s}{s+1} \log_F\left(\max\left(\frac{enF^{1/s}}{\lambda_t}, 1\right)\right).
\end{align*}
\end{definition}

The definition of $h(\lambda)$ in this case is used as a
% The potential function is composed of two main terms, the fitness and a
penalty term that grows linearly in $\log_F \lambda$ (since $-\log_F\left(\frac{enF^{1/s}}{\lambda_t}\right) = -\log_F(enF^{1/s}) + \log_F(\lambda_t)$). That is, when $\lambda$ increases the penalty decreases and vice-versa.
The idea behind this definition is that small values of $\lambda$ may lead to decreases in fitness, but these are compensated by an increase in $\lambda$ and a reduction of the penalty term.

Since the range of the penalty term is limited, the potential is always close to the current fitness as shown in the following lemma.

\begin{lemma}\label{lem:relation-fitness-potential}
For all generations~$t$, the fitness and the potential are related as follows: $f(x_t) - \frac{2s}{s+1} \log_F(enF^{1/s}) \le g_1(X_t) \le f(x_t)$. In particular, $g_1(X_t)=n$ implies ${f(x_t)=n}$.
\end{lemma}
\begin{proof}
The penalty term $\frac{2s}{s+1}\log_F\left(\max\left(\frac{en F^{1/s}}{\lambda_t}, 1\right)\right)$ is a non-increasing function in $\lambda_t$ with its minimum being $0$ for $\lambda\ge en F^{1/s}$ and its maximum being $\frac{2s}{s+1}\log_F\left(en F^{1/s}\right)$ when $\lambda=1$. Hence, $f(x_t) - \frac{2s}{s+1} \log_F(enF^{1/s}) \le g_1(X_t) \le f(x_t)$.
% The penalty term $\frac{2s}{s+1}\log_F\left(\frac{e F^{1/s}n^3}{\lambda_t}\right)$ is a non-increasing function in $\lambda_t$ with its maximum being $\frac{2s}{s+1}\log_F\left(e F^{1/s}n^3\right)$ when $\lambda=1$. By Lemma~\ref{lem:probability-exceed-polyn} $\lambda_t\le eF^{1/s}n^3$ for all $t$ with probability $1-\exp{(\Omega(n^2))}$. With the same probability the penalty term is at least $0$. Hence, $f(x_t) - \frac{2s}{s+1} \log_F(eF^{1/s}n^3) \le g_1(X_t) \le f(x_t)$ with probability $1-\exp{(\Omega(n^2))}$ .
\end{proof}

Now we proceed to show that with the correct choice of hyper-parameters the drift in potential is at least a positive constant during all parts of the optimisation.

\begin{lemma}\label{lem:potentialDrift}
Consider the \saocl as in Theorem~\ref{thm:optimisationTime-generations}. Then for every generation $t$ with $f(x_t) < n$,
\[
    \E{g_1(X_{t+1})-g_1(X_{t})\mid X_{t}} \ge \frac{1-s}{2e}.
\]
for large enough $n$.
This also holds when only considering improvements that increase the fitness by~$1$.
\end{lemma}
\begin{proof}
Given that $h(\lambda_t)=-\frac{2s}{s+1} \log_F\left(\max\left(\frac{enF^{1/s}}{\lambda_t},1\right)\right)$ is a non-decreasing function, if $\lambda\le F$ then $h(1)\ge h(\lambda/F)$.
% \begin{align*}
%     h(1)
%     &= -\frac{2s}{s+1}\left(\log_F(eF^{1/s}n^3)\right)\\
%     &\ge -\frac{2s}{s+1}\left(\log_F(eF^{1/s}n^3)\right) - \frac{2s}{s+1} (1-\log_F(\lambda))\\
%     &=-\frac{2s}{s+1}\left(\log_F(eF^{1/s}n^3)\right) + \frac{2s}{s+1}\log_F\left(\frac{\lambda}{F}\right)\\
%     &=-\frac{2s}{s+1}\left(\log_F\left(\frac{eF^{1/s}n^3}{\lambda/F}\right)\right)\\
%     &= h(\lambda/F)
% \end{align*}
% \begin{align*}
%     h(1)- h(\lambda F^{1/s})
%     &=-\frac{2s}{s+1}\left(\log_F(\lambda F^{1/s})\right)
%     = -\frac{2s}{s+1}\left(\log_F(\lambda)+\frac{1}{s}\right) \\
%     &\ge -2
%     = -\frac{2s}{s+1}\left(\frac{s+1}{s}\right)
%     = \frac{2s}{s+1}\left(\log_F\left(\frac{\lambda}{F}\right)-\log_F\left(\lambda F^{1/s}\right)\right) \\
%     &=  h(\lambda/F)- h(\lambda F^{1/s})
% \end{align*}
Hence, by Lemma~\ref{lem:generalised-potential}, for all $\lambda$, $\E{g_1(X_{t+1})-g_1(X_{t})\mid X_{t}}$ is at least
\begin{align}
     &\left(\Deltagain+h(\lambda/F)-h(\lambda F^{1/s})\right)\pimp + h(\lambda F^{1/s}) -h(\lambda) -\Deltaloss \ploss . \label{eq:potential-with-h}
\end{align}
We first consider the case $\lambda_t\le en$ as then $\lambda_{t+1}\le enF^{1/s}$ and $h(\lambda_{t+1})
= - \frac{2s}{s+1} (\log_F(enF^{1/s}) - \log_F(\lambda_{t+1}))<0$. Hence, $\E{g_1(X_{t+1}) - g_1(X_t) \mid X_t, \lambda_t \le en}$ is at least
\begin{align*}
     &\left(\Deltagain+\frac{2s}{s+1} \log_F\left(\frac{\lambda}{F}\right)-\frac{2s}{s+1} \log_F\left(\lambda F^{1/s}\right)\right)\pimp+\frac{2s}{s+1} \log_F\left(\lambda F^{1/s}\right)\\* %no page break here
     &\hspace{7cm}-\frac{2s}{s+1} \log_F\left(\lambda\right)-\Deltaloss\ploss\\
     =\;& \left(\Deltagain-\frac{2s}{s+1}\left(\frac{s+1}{s}\right)\right)\pimp +\frac{2s}{s+1}\left(\frac{1}{s}\right)-\Deltaloss\ploss\\
     =\;& \frac{2}{s+1}+\left(\Deltagain-2\right)\pimp-\Deltaloss\ploss .
\end{align*}
By Lemma~\ref{lem:bounds_probabilites} $\Deltagain\ge1$, hence $\E{g_1(X_{t+1}) - g_1(X_t) \mid X_t, \lambda_t \le en}\ge \frac{2}{s+1}-\pimp-\Deltaloss\ploss$.
Using $\frac{2}{s+1}=\frac{s+1+1-s}{s+1}
=1+\frac{1-s}{s+1}$ yields
\begin{align*}
    \E{g_1(X_{t+1}) - g_1(X_t) \mid X_t, \lambda_t \le en}&\ge 1+\frac{1-s}{s+1}-\pimp-\Deltaloss\ploss%\\
    %&=\frac{1-s}{s+1}+(1-p_{i,1}^+)^{\round{\lambda}}-\Deltaloss(p_{i,1}^-)^{\round{\lambda}}.
\end{align*}
By Lemma~\ref{lem:bounds_probabilites} this is at least
\begin{align}
    &\frac{1-s}{s+1}+\left(1-1.14\left(\frac{n-i}{n}\right)\left(1-\frac{1}{n}\right)^{n-1}\right)^{\round{\lambda}}\mkern-25mu-\left(\frac{e}{e-1}\right)\left(1-\frac{n-i}{en}-\left(1-\frac{1}{n}\right)^n\right)^{\round{\lambda}}\nonumber\\
    &\ge\frac{1-s}{s+1}+\left(1-\frac{1.14}{e\left(1-\frac{1}{n}\right)}\left(\frac{n-i}{n}\right)\right)^{\round{\lambda}}-\left(\frac{e}{e-1}\right)\left(1-\frac{n-i}{en}-\frac{1}{e}\left(1-\frac{1}{n}\right)\right)^{\round{\lambda}}\nonumber\\
    &=\frac{1-s}{s+1}+\left(1-\frac{1.14}{e}\left(\frac{n-i}{n-1}\right)\right)^{\round{\lambda}}-\left(\frac{e}{e-1}\right)\left(\frac{e-1}{e}-\frac{n-i-1}{en}\right)^{\round{\lambda}}\label{eq:potential_with_general_lambda}.
\end{align}
We start taking into account only $\round{\lambda}\ge2$, that is, $\lambda\ge1.5$  and  later  on  we  will  deal with $\round{\lambda}=1$. For $\round{\lambda} \ge 2$, $\E{g_1(X_{t+1}) - g_1(X_t) \mid X_t, 1.5 \le \lambda_t \le en}$ is at least
\begin{align*}
    &\frac{1-s}{s+1}+\left(1-\frac{1.14}{e}\left(\frac{n-i}{n-1}\right)\right)^{\round{\lambda}}-\left(\frac{e}{e-1}\right)^{\round{\lambda}/2}\left(\frac{e-1}{e}-\frac{n-i-1}{en}\right)^{\round{\lambda}}\\
    &=\frac{1-s}{s+1}+\left.\underbrace{\left(1-\frac{1.14}{e}\left(\frac{n-i}{n-1}\right)\right)}_{y_1}\right.^{\round{\lambda}}-\left.\underbrace{\left(\left(\frac{e-1}{e}\right)^{1/2}-\frac{n-i-1}{(e^2-e)^{1/2}n}\right)}_{y_2}\right.^{\round{\lambda}}
\end{align*}
Let $y_1$ and $y_2$ be the respective bases of the terms raised to $\round{\lambda}$ as indicated above. We will now prove that $y_1\ge y_2$ for all $0\le i<n$ which implies that $\E{g_1(X_{t+1}) - g_1(X_t) \mid X_t, 1.5 \le \lambda_t \le en}\ge \frac{1-s}{s+1}\ge\frac{1-s}{2e}$, where the last inequality holds because $s<1$.

The terms $y_1$ and $y_2$ can be described by linear equations ${y_1=m_1 (n-i)+b_1}$ and ${y_2=m_2 (n-i)+b_2}$ with $m_1=-\frac{1.14}{e(n-1)}$, $b_1=1$, $m_2=-\frac{1}{n\sqrt{e^2-e}}$ and ${b_2=\sqrt{\frac{e-1}{e}}+\frac{1}{n\sqrt{e^2-e}}}$.
Since $m_2<m_1$ for all $n \ge 11$, the difference $y_1 - y_2$ is minimised for $n-i=1$. When $n-i=1$, then $y_1=1-\frac{1.14}{e(n-1)}>\left(\frac{e-1}{e}\right)^{1/2}=y_2$ for all $n>3$, therefore $y_1>y_2$ for all $0\le i<n$.

When $\round{\lambda}=1$, from Equation~\eqref{eq:potential_with_general_lambda} $\E{g_1(X_{t+1}) - g_1(X_t) \mid X_t, \lambda_t \le 1.5}\ge \frac{1-s}{s+1}-\frac{1.14}{e}\left(\frac{n-i}{n-1}\right)+\frac{n-i-1}{(e-1)n}$
which is monotonically decreasing for $0<i<n-1$ when ${n>e/(1.14-0.14e)\approx3.58}$, hence $\E{g_1(X_{t+1}) - g_1(X_t) \mid X_t, \lambda_t \le 1.5}\ge\frac{1-s}{s+1}-\frac{1.14}{e(n-1)}$ which is bounded by $\frac{1-s}{2e}$ for large enough $n$ since $s<1$.

Finally, for the case $\lambda_t>en$, in an unsuccessful generation the penalty term is capped, hence $h(\lambda F^{1/s})=h(\lambda)$. Then by Equation \eqref{eq:potential-with-h}, $\E{g_1(X_{t+1}) - g_1(X_t) \mid X_t, \lambda_t > en}$ is at least
\begin{align*}
     &\left(\Deltagain+h(\lambda/F)-h(\lambda)\right)\pimp -\Deltaloss \ploss\\
     &=\left(\Deltagain+\frac{2s}{s+1} \log_F\left(\frac{\lambda}{F}\right)-\frac{2s}{s+1} \log_F\left(\lambda\right)\right)\pimp -\Deltaloss \ploss\\
     &=\left(\Deltagain-\frac{2s}{s+1}\right)\pimp -\Deltaloss \ploss
\end{align*}
By Lemma~\ref{lem:bounds_probabilites}, $\lambda_t > en$ implies
$\pimp \ge 1-\left(1-\frac{1}{en}\right)^{en} \ge 1 - \frac{1}{e}$
and $\ploss \Deltaloss \le \left(\frac{e-1}{e}\right)^{en} \frac{e}{e-1} = \left(\frac{e-1}{e}\right)^{en-1}$.
Together,
\begin{align*}
    & \E{g_1(X_{t+1}) - g_1(X_t) \mid X_t, \lambda_t > en}\\
    & \ge\; \left(\Deltagain - \frac{2s}{s+1}\right)\pimp  - \Deltaloss\ploss\\
    & \ge\; \left(1-\frac{1}{e}\right)\left(1-\frac{2s}{s+1}\right) - \left(\frac{e-1}{e}\right)^{en-1}\\
    & =\; \left(\frac{1}{e} + \left(1-\frac{2}{e}\right)\right)\left(1-\frac{2s}{s+1}\right) - \left(\frac{e-1}{e}\right)^{en-1}\\
    & =\;\frac{1}{e} \left(1-\frac{2s}{s+1}\right) +  \left(1-\frac{2}{e}\right)\left(1-\frac{2s}{s+1}\right) - \left(\frac{e-1}{e}\right)^{en-1} .
\end{align*}
The term 
%$\left(1 - \frac{1}{e}\right) = \frac{1}{e} + \left(1-\frac{2}{e}\right)$ and
$\left(1-\frac{2}{e}\right)\left(1 - \frac{2s}{s+1}\right)$ is a positive constant, hence, for large enough~$n$ this term is larger than $\left(\frac{e-1}{e}\right)^{en-1}$ and
\[
    \E{g_1(X_{t+1}) - g_1(X_t) \mid X_t, \lambda_t > en} \ge \frac{1}{e} \left(1 - \frac{2s}{s+1}\right) = \frac{1}{e} \left(\frac{1-s}{s+1}\right) \ge
    \frac{1-s}{2e}.
\]
Since $s < 1$, this is a strictly positive constant.
\end{proof}

With this constant lower bound on the drift of the potential, the proof of  Theorem~\ref{thm:optimisationTime-generations} is now quite straightforward.
\begin{proof}[Proof of Theorem~\ref{thm:optimisationTime-generations}]
We bound the time to get to the optimum using the potential function $g_1(X_t)$. Lemma~\ref{lem:potentialDrift} shows that the potential has a positive constant drift whenever the optimum has not been found, and by Lemma~\ref{lem:relation-fitness-potential} if $g_1(X_t)=n$ then the optimum has been found. Therefore, we can bound the number of generations by the time it takes for $g_1(X_t)$ to reach $n$.

To fit the perspective of the additive drift theorem (Theorem~\ref{thm:additive_drift}) we switch to the function ${\overline{g_1}(X_t) := n-g_1(X_t)}$ and note that $\overline{g_1}(X_t)=0$ implies that ${g_1(X_t) = f(x_t) = n}$. The initial value $\overline{g_1}(X_0)$ is at most
$n + \frac{2s}{s+1}\log_F\left(e nF^{1/s}\right)$ by Lemma~\ref{lem:relation-fitness-potential}. Using Lemma~\ref{lem:potentialDrift} and the additive drift theorem, the expected number of generations is at most
\begin{align*}
    \frac{n+\frac{2s}{s+1}\log_F\left(e nF^{1/s}\right)}{ \frac{1-s}{2e}}=O(n).
\end{align*}
\end{proof}
\subsection{Bounding the Number of Evaluations}
\label{sec:evaluations}

A bound on the number of generations, by itself, is not sufficient to claim that the \saocl is efficient in terms of the number of evaluations. Obviously, the number of evaluations in generation~$t$ equals $\lambda_t$ and this quantity is being self-adjusting over time. So we have to study the dynamics of $\lambda_t$ more carefully. Since $\lambda$ grows exponentially in unsuccessful generations, it could quickly attain very large values. However, we show that this is not the case and only $O(n \log n)$ evaluations are sufficient, in expectation.
\begin{theorem}
\label{thm:optimisationTime}
Let the update strength $F > 1$ and the success rate $0 < s < 1$ be constants.
The expected number of function evaluations of the \saocl on  \onemax is $O(n\log n)$.
\end{theorem}

Bounding the number of evaluations is more challenging than bounding the number of generations as we need to keep track of the offspring population size~$\lambda$ and how it develops over time. Large values of $\lambda$ lead to a large number of evaluations made in one generation. Small values of $\lambda$ can lead to a fallback.

In the elitist \saopl, small values of $\lambda$ are not an issue since there are no fallbacks. The analysis of the \saopl in Section~\ref{sec:elitist-saopl} relies on every fitness level being visited at most once. In our non-elitist algorithm, this is not guaranteed. Small values of $\lambda$ can lead to decreases in fitness, and then the same fitness level can be visited multiple times.

The reader may think that small values of $\lambda$ only incur few evaluations and that the additional cost for a fallback is easily accounted for. However, it is not that simple. Imagine a fitness level~$i$ and a large value of $\lambda$ such that a fallback is unlikely. But it is possible for $\lambda$ to decrease in a sequence of improving steps. Then we would have a small value of $\lambda$ and possibly a sequence of fitness-decreasing steps. Suppose the fitness decreases to a value at most~$i$, then if $\lambda$ returns to a large value, we may have visited fitness level~$i$ multiple times, with large (and costly) values of~$\lambda$.

It is possible to show that, for sufficiently challenging fitness levels, $\lambda$ moves towards an equilibrium state, i.\,e.\ when $\lambda$ is too small, it tends to increase. However, this is generally not enough to exclude drops in $\lambda$. Since $\lambda$ is multiplied or divided by a constant factor in each step, a sequence of $k$ improving steps decreases $\lambda$ by a factor of $F^k$, which is exponential in~$k$. For instance, a value of $\lambda = \log^{O(1)}n$ can decrease to $\lambda = \Theta(1)$ in only $O(\log \log n)$ generations. We found that standard techniques such as the negative drift theorem, applied to $\log_F(\lambda_t)$, are not strong enough to exclude drops in $\lambda$.

We solve this problem as follows. We consider the best-so-far fitness $f_t^* = \max\{f(x_{t'}) \mid 0 \le t' \le t\}$ at time~$t$ (as a theoretical concept, as the \saocl is non-elitist and unaware of the best-so-far fitness). We then divide the run into fitness intervals of size $\log n$ that we call \emph{blocks}, and bound the time for the best-so-far fitness to reach a better block. To this end, we reconsider the potential function used to bound the expected number of generations in Theorem~\ref{thm:optimisationTime-generations} and refine our arguments to obtain a bound on the expected number of \emph{generations} to increase the best-so-far fitness by $\log n$ (see Lemma~\ref{lem:generations-for-b-a-block} below). Denoting by $b$ the target fitness of a better block, in the current block the fitness is at most $b-1$. To bound the number of \emph{evaluations}, we show that the offspring population size is likely to remain in $O(1/p_{b-1, 1}^+)$, where $p_{b-1, 1}^+$ is the worst-case improvement probability for a single offspring creation in the current block. An application of Wald's equation bounds the total expected number of evaluations in all generations until a new block is reached.

At the time a new block $i$ is reached, the current offspring population size $\lambda^{(i)}$ is not known, yet it contributes to the expected number of evaluations during the new block. We provide tail bounds on $\lambda^{(i)}$ to show that excessively large values of $\lambda^{(i)}$ are unlikely. This way we bound the total contribution of $\lambda^{(i)}$'s across all blocks~$i$ by $O(n \log n)$.

\begin{lemma}
\label{lem:generations-for-b-a-block}
Consider the \saocl as in Theorem~\ref{thm:optimisationTime}.
For every $a, b\in \{0, \dots, n\}$, the expected number of generations to increase the current fitness from a value at least~$a$ to at least~$b > a$ is at most
\[
    \frac{b-a + \frac{2s}{s+1}\log_F\left(e nF^{1/s}\right)}{\frac{1-s}{2e}} = O(b-a + \log n).
\]
For $b = a + \log n$, this bound is $O(\log n)$.
\end{lemma}
\begin{proof}
We use the proof of Theorem~\ref{thm:optimisationTime-generations} with a revised potential function of ${\overline{g_1}'(X_t) := \max(\overline{g_1}(X_t) - (n-b), 0)}$ and stopping when $\overline{g_1}'(X_t)=0$ (which implies that a fitness of at least $b$ is reached) or a fitness of at least $b$ is reached beforehand. Note that the maximum caps the effect of fitness improvements that jump to fitness values larger than~$b$. As remarked in Lemma~\ref{lem:potentialDrift}, the drift bound for $g_1(X_t)$ still holds when only considering fitness improvements by~1. Hence, it also holds for $\overline{g_1}'(X_t)$ and the analysis goes through as before.
\end{proof}

In our preliminary publication~\cite{Hevia2021} we introduced a novel analysis tool that we called \emph{ratchet argument}. We considered the best-so-far fitness $f_t^* = \max\{f(x_{t'}) \mid 0 \le t' \le t\}$ at time~$t$ (as a theoretical concept, as the \saocl is non-elitist and unaware of the best-so-far fitness) and used drift analysis to show that, with high probability, the current fitness never drops far below $f_t^*$, that is, $f(x_t) \ge f_t^* - r \log n$ for a constant~$r > 0$.
We called this a \emph{ratchet argument}\footnote{This name is inspired by the term ``Muller's ratchet'' from biology~\cite{Felsenstein1974} that considers a ratchet mechanism in asexual evolution, albeit in a different context.}: if the best-so-far fitness increases, the lower bound on the current fitness increases as well. The lower bound thus works like a ratchet mechanism that can only move in one direction.
Our revised analysis no longer requires this argument. We still present the following lemma since (1) it might be of interest as a structural result about the typical behaviour of the algorithm, (2) it has found applications in follow-up work~\cite{Hevia2021FOGA} of~\cite{Hevia2021} and it makes sense to include it here for completeness and (3) the basic argument may prove useful in analysing other non-elitist algorithms.
%
%The following lemma presents our ratchet argument and 
Lemma~\ref{lem:global-failures} also shows that with high probability the fitness does not decrease when $\lambda \ge 4 \log n$. 
A proof is given in the appendix.
% \begin{lemma}
% \label{lem:global-failures}
% Consider the \saocl as in Theorem~\ref{thm:optimisationTime}.
% Let ${f^*_t := \max_{t' \le t} f(x_{t'})}$ be its best-so-far fitness at generation~$t$ and let $T$ be the first generation in which the optimum is found.
% Then with probability $1-O(1/n)$ the following statements hold for a large enough constant~$r > 0$ (that may depend on~$s$).
% \begin{enumerate}
%     \item For all~$t \le T$ in which $\lambda_t \ge 4\log n$, we have $f(x_{t+1}) \ge f(x_t)$.
%     \item For all~$t \le T$, the fitness is at least: $f(x_t) \ge f^*_t - r\log n$.
% \end{enumerate}
% \end{lemma}

%\dirk{Not sure whether it's a good idea to keep Lemma~\ref{lem:global-failures}.}

\textEnd{This appendix contains the proof of Lemma~\ref{lem:global-failures} that was omitted from the main part.}
\begin{theoremEnd}[restate, no link to proof]{lemma}
\label{lem:global-failures}
Consider the \saocl as in Theorem~\ref{thm:optimisationTime}.
Let ${f^*_t := \max_{t' \le t} f(x_{t'})}$ be its best-so-far fitness at generation~$t$ and let $T$ be the first generation in which the optimum is found.
Then with probability $1-O(1/n)$ the following statements hold for a large enough constant~$r > 0$ (that may depend on~$s$).
\begin{enumerate}
    \item For all~$t \le T$ in which $\lambda_t \ge 4\log n$, we have $f(x_{t+1}) \ge f(x_t)$.
    \item For all~$t \le T$, the fitness is at least: $f(x_t) \ge f^*_t - r\log n$.
\end{enumerate}
\end{theoremEnd}

\begin{proofEnd}
Let $E_1^t$ denote the event that $\lambda_t < 4 \log n$ or $f(x_{t+1}) \ge f(x_t)$.
Hence we only need to consider $\lambda_t$-values of $\lambda_t \ge 4 \log n \ge 2\log_{\frac{e}{e-1}}n$ and by Lemma~\ref{lem:bounds_probabilites} Equation~\eqref{eq:ploss} we have
\[
    \Prob{\overline{E_1^t}} \le \left(\frac{e-1}{e}\right)^{\lambda_t}
    \le \left(\frac{e-1}{e}\right)^{2\log_{\frac{e}{e-1}}n}
    = \frac{1}{n^2}.
\]
Given that the event $\overline{E_1^t}$ happens in each step with probability at most~$\frac{1}{n^2}$, by a union bound, the probability that this happens in the first $T$ generations, with $T$ being a random variable with $\E{T}<\infty$, is at most ${\sum_{t=1}^\infty \Prob{T = t} \cdot t/n^2} = \E{T}/n^2$, and by Theorem~\ref{thm:optimisationTime-generations} $\E{T}/n^2=O(1/n)$. Hence, the probability that the first statement holds is $1-O(1/n)$.
For the second statement, let $t^*$ be a generation in which the best-so-far fitness was attained: $f(x_{t^*}) = f^*_t$. By Lemma~\ref{lem:relation-fitness-potential}, abbreviating $\alpha := \frac{2s}{s+1} \log_F(enF^{1/s})$, the condition $f(x_{t^*}) \ge f(x_t) + r \log n$ implies
$
    g_1(X_{t^*}) \ge f(x_{t^*}) - \alpha \ge f(x_t) - \alpha + r \log n \ge g_1(X_t) - \alpha + r \log n$.

Now define events
$
    E_2^t = (\forall t' \in [t+1, n^2] \colon g_1(X_{t'}) \ge g_1(X_t) + \alpha - r \log(n))$.
We apply the negative drift theorem (Theorem~\ref{thm:negative-drift}) to bound $\Prob{\overline{E_2^t}}$ from above.
For any $t < n^2$ let $a := g_1(X_t) - r\log n + \alpha$ and $b := g_1(X_t) < n$, where $r > \alpha$ will be chosen later on. We pessimistically assume that the fitness component of $g$ can only increase by at most~1. Lemma~\ref{lem:potentialDrift} has already shown that, even under this assumption, the drift is at least a positive constant.
This implies the first condition of Theorem~2 in~\cite{Oliveto2012Erratum}. For the second condition, we need to bound transition probabilities for the potential. Owing to our pessimistic assumption, the current fitness can only increase by at most~1.
The fitness only decreases by~$j$ if \emph{all} offspring are worse than their parent by at least~$j$. Hence, for all $\lambda$, the decrease in fitness is bounded by the decrease in fitness of the \emph{first} offspring.
The probability of the first offspring decreasing fitness by at least~$j$ is bounded by the probability that $j$ bits flip, which is in turn bounded by $1/(j!) \le 2/2^j$. The possible penalty in the definition of $g$ changes by at most $\max\left(\frac{se}{e-1}, \frac{se}{e-1} \cdot \frac{1}{s}\right) = \frac{e}{e-1} < 1$. Hence, for all~$t$,
\[
    \Prob{\lvert g_1(X_{t-1})-g_1(X_t)\rvert \ge j+1 \mid g_1(X_t) > a} \le \frac{4}{2^{j+1}},
\]
which meets the second condition of Theorem~2 in~\cite{Oliveto2012Erratum}. It then states that there is a constant $c^*$ such that the probability that within $2^{c^*(a-b)/4}$ generations a potential of at most~$a$ is reached, starting from a value of at least~$b$, is $2^{-\Omega(a-b)}$. By choosing the constant~$r$ large enough, we can scale up $a-b$ and thus make $2^{c^*(a-b)/4} \ge n^2$ and $2^{-\Omega(a-b)} = O(1/n^2)$.
This yields $\Prob{\overline{E_2^t}} = O(1/n^2)$.

Arguing as before, using a union bound we show that the probability that $\overline{E_2^t}$ happens during the first $T$ generations is at most $\sum_{t=1}^\infty \Prob{T = t} \cdot t\cdot O(1/n^2) = O(1/n)$.
% $O(n)$ expected generations is $O(1/n)$. 
By Markov's inequality, the probability of not finding the optimum in $n^2$ generations, that is, $\prob{T\ge n^2}$ is at most $\E{T}/n^2=O(1/n)$ as well. Adding up all failure probabilities completes the proof.
\end{proofEnd}

% Lemma~\ref{lem:global-failures} shows that even though the algorithm is non-elitist if the current offspring population size is sufficiently large the algorithm typically does not reduce its current fitness. Additionally, it shows that throughout the optimisation the algorithm maintains a fitness that is at most a logarithmic factor smaller than the best-so-far fitness with high probability. 

In~\cite{Hevia2021} we divided the optimisation in blocks of length $\log n$ and with the help of the ratchet argument shown in Lemma~\ref{lem:global-failures} and other helper lemmas we showed that each block is typically optimised efficiently. Adding the time spent in each block, we obtained that the algorithm optimises \onemax in $O(n\log n)$ evaluations with high probability. 
It is straightforward to derive a bound on the number of expected evaluations of the same order.
%Since the probability that the algorithm deviates from the typical behaviour described in Lemma~\ref{lem:global-failures} is small, the expected runtime of the algorithm is asymptotically the same as a typical run. 

In this revised analysis we still divide the optimisation in blocks of length $\log n$, but use simpler and more elegant arguments to compute the time spent in a block and the total expected runtime. To bound the time spent optimising a block, first we divide each block on smaller chunks called \emph{phases} and bound the time spent in each phase. This is shown in the following lemma.

\begin{lemma}
\label{lem:bound-on-expected-lambda-t-v2}
Consider the \saocl as in Theorem~\ref{thm:optimisationTime}.
Fix a fitness value $b$ and denote the current offspring population size by $\lambda_0$.
%in generation~$t^*$.
Define $\lambdabar := CF^{1/s}/p_{b-1, 1}^+$ for a constant $C > 0$  that may depend on $F$ and $s$ that
satisfies 
\begin{equation}
\label{eq:condition-for-C}
    \left(\frac{s+1}{s} \cdot e^{1-C}\right)^{\frac{s}{s+1}} \le \frac{F^{-1/s}}{2}.
\end{equation}
Define a \emph{phase} as a sequence of generations that ends in the first generation where $\lambda$ attains a value of at most~$\lambdabar$ or a fitness of at least~$b$ is reached.
Then the expected number of evaluations made in that phase is $O(\lambda_0)$.

%The same statement holds if $\lambdabar$ is redefined to a time-dependent variable $\lambdabar_t := CF^{1/s}/p_{b^*_t,1}^+$ where $b_t^*$ denotes the best-so-far fitness at time~$t$.
%If the phase ends because a fitness larger than~$b$ is reached
\end{lemma}
We note that a constant $C > 0$ meeting inequality~\eqref{eq:condition-for-C} exists since the left-hand side converges to~0 when $C$ goes to infinity, while the right-hand side remains a positive constant.
\begin{proof}[Proof of Lemma~\ref{lem:bound-on-expected-lambda-t-v2}]
%W.\,l.\,o.\,g.\ we assume that the generation index at the start of the phase is $0$. 
If $\lambda_0 F^{1/s} \le \lambdabar$ or if the current fitness is at least~$b$ then the phase takes only one generation and $\lambda_0$ evaluations as claimed. Hence we assume in the following that $\lambdabar F^{-1/s} < \lambda_0$ and that the current fitness is less than~$b$.

We use some ideas from the proof of Theorem~9 in~\cite{Doerr2018}\footnote{Said theorem only holds for values of $F > 1$ that can be chosen arbitrarily small. We generalise the proof to work for arbitrary constant~$F$. This requires the use of a stronger Chernoff bound.} that bounds the expected number of evaluations in the self-adjusting (1+($\lambda$,$\lambda$))~GA.
Let $Z$ denote the random number of iterations in the phase and let $T$ denote the random number of evaluations in the phase. 
Since $\lambda_t$ can only grow by $F^{1/s}$, $\lambda_{i} \le \lambda_{0} \cdot F^{i/s}$ for all $i \in \N_0$.
If $Z = z$, the number of evaluations is bounded by 
\[
    \E{T \mid Z = z} \le \lambda_{0} \cdot \sum_{i=1}^{z} F^{i/s} = \lambda_{0} \cdot \frac{F^{\frac{z+1}{s}}-F^{1/s}}{F^{1/s}-1} 
    \le \lambda_{0} \cdot \frac{F^\frac{z+1}{s}}{F^{1/s}-1}.
\]
While $\lambda_t \ge \lambdabar F^{-1/s}$ and the current fitness is $i < b$, the probability of an improvement is at least
\[
    1 - \left(1 - p_{i, 1}^+\right)^{\lambda_t} \ge 
    1 - \left(1 - p_{b-1, 1}^+\right)^{\lambdabar F^{-1/s}} \ge 1 - e^{-\lambdabar F^{-1/s}\cdot p_{b-1, 1}^+} = 1 - e^{-C}.
\]
%Now $\lambda_z \le \lambda_0 \le \lambdabar$ if there are at most $z \cdot \frac{s}{s+1}$ unsuccessful iterations since then there are at most $z \cdot \frac{s}{s+1}$ iterations increasing $\log_F(\lambda)$  by $1/s$ each, and at least $z \cdot \frac{1}{s+1}$ iterations decreasing $\log_F(\lambda)$ by $1$ each. 
If during the first $z$ iterations we have at most $z \cdot \frac{s}{s+1}$ unsuccessful iterations, this implies that at least $z \cdot \frac{1}{s+1}$ iterations are successful. The former steps increase $\log_F(\lambda)$ 
%at least $z \cdot \frac{s}{s+1}$ times 
by $1/s$ each, and the latter steps decrease $\log_F(\lambda)$ by $1$ each.
%in each of at least $z \cdot \frac{1}{s+1}$ steps.
In total, we get $\lambda_z \le \lambda_0 \cdot (F^{1/s})^{z \cdot s/(s+1)} \cdot (1/F)^{Z/(s+1)} = \lambda_0$ and thus $\lambda_z \le \lambda_0 \le \lambdabar$. We conclude that having at most $z \cdot \frac{s}{s+1}$ unsuccessful iterations among the first $z$ iterations is a sufficient condition for ending the phase within $z$ iterations.

We define independent random variables $\{Y_t\}_{t \ge t^*}$ such that $Y_t \in \{0, 1\}$, $\Prob{Y_t = 0} = 1-e^{-C}$ and $\Prob{Y_t = 1} = e^{-C}$. Denote $Y := \sum_{i=1}^z Y_i$ and note that $\E{Y} = z \cdot e^{-C}$.
Using classical Chernoff bounds (see, e.\,g.\ Theorem~10.1 in~\cite{DoerrProbabilityChapter2020}),
\begin{align*}
    \Prob{Z = z} \le \Prob{Z \ge z} \le\;& \Prob{Y \ge z \cdot \frac{s}{s+1}}\\
    =\;& \Prob{Y \ge \E{Y} \cdot \left(\frac{s}{s+1} \cdot e^C\right)}\\
    \le\;& \left(\frac{e^{\frac{s}{s+1} \cdot e^{C} - 1}}{\left(\frac{s}{s+1} \cdot e^C\right)^{\frac{s}{s+1} \cdot e^C}}\right)^{z \cdot e^{-C}}\\
    \le\;& \left(\frac{e^{\frac{s}{s+1} \cdot e^{C}}}{\left(\frac{s}{s+1} \cdot e^C\right)^{\frac{s}{s+1} \cdot e^C}}\right)^{z \cdot e^{-C}}
    %=\;& \left(\frac{e^{\frac{s}{s+1}}}{\left(\frac{s}{s+1} \cdot e^C\right)^{\frac{s}{s+1}}}\right)^{z}\\
    % =\;& \left(\frac{e^{\frac{s}{s+1}}}{\left(\frac{s}{s+1} \cdot e^C\right)^{\frac{s}{s+1}}}\right)^{z}\\
    % =\;& \left(\frac{e^{\frac{s}{s+1} - e^{-C}}}{\left(\frac{s}{s+1} \cdot e^C\right)^{\frac{s}{s+1}}}\right)^{z}\\
    % =\;& \left(\left(\frac{s+1}{s} \cdot e^{-C} \cdot (e^{1 - \frac{s+1}{s} \cdot e^{-C}})\right)^{\frac{s}{s+1}}\right)^{z}\\
    = \left(\left(\frac{s+1}{s} \cdot e^{1-C}\right)^{\frac{s}{s+1}}\right)^{z}.
\end{align*}
By assumption on~$C$ this is at most $(F^{-1/s}/2)^z = F^{-z/s} \cdot 2^{-z}$. 

Putting things together,
\begin{align*}
    \E{T} =\;& \sum_{z=1}^{\infty} \Prob{Z = z} \cdot \E{T \mid Z = z}\\
%    \le\;& \sum_{z=1}^{\infty} \Prob{Z \ge z} \cdot \lambda_0 \cdot \frac{F^{\frac{z+1}{s}}}{F^{1/s}-1}\\
    \le\;& \sum_{z=1}^{\infty} F^{-z/s} \cdot 2^{-z} \cdot \lambda_0 \cdot \frac{F^{\frac{z+1}{s}}}{F^{1/s}-1}\\
    =\;& \lambda_0 \cdot \frac{F^{1/s}}{F^{1/s}-1} \cdot \sum_{z=1}^{\infty} 2^{-z}
    = \lambda_0 \cdot \frac{F^{1/s}}{F^{1/s}-1}.
\end{align*}
%The last statement holds since $\lambdabar_t$ is non-decreasing and the estimations of transition probabilities remain valid.
\end{proof}

%\subsubsection{Analysing the Number of Evaluations in a Block}

We now use Lemma~\ref{lem:bound-on-expected-lambda-t-v2} and Wald's equation to compute the expected number of evaluations spent in a block.

\begin{lemma}
\label{lem:bound-on-expected-evals-to-increase-fitness}
Consider the \saocl as in Theorem~\ref{thm:optimisationTime}.
Starting with a fitness of~$a$ and an offspring population size of~$\lambda_0$, the expected number of function evaluations until a fitness of at least~$b$ is reached for the first time is at most
\[
    \E{\lambda_0 + \dots + \lambda_t \mid \lambda_0} \le O(\lambda_0) + O(b-a + \log n) \cdot \frac{1}{p_{b-1, 1}^+}.
    %\cdot \left(F^{1/s} + \frac{F^{1/s}}{\ln F}\right).
\]
\end{lemma}
\begin{proof}
We use the variable $\lambdabar$ and the definition of a \emph{phase} from the statement of Lemma~\ref{lem:bound-on-expected-lambda-t-v2}.
In the first phase, the number of evaluations is bounded by $O(\lambda_0)$ by Lemma~\ref{lem:bound-on-expected-lambda-t-v2}.
Afterwards, we either have a fitness of at least~$b$ or a $\lambda$-value of at most~$\lambdabar$. In the former case we are done. In the latter case, we apply Lemma~\ref{lem:bound-on-expected-lambda-t-v2} repeatedly until a fitness of at least~$b$ is reached. In every considered phase the expected number of evaluations is at most $O(\lambdabar) = O(1/p_{b-1, 1}^+)$. Note that all these applications of Lemma~\ref{lem:bound-on-expected-lambda-t-v2} yield a bound that is irrespective of the current fitness and the current offspring population size. Hence these upper bounds can be thought of as independent and identically distributed random variables. 

By Lemma~\ref{lem:generations-for-b-a-block} the expected number of generations to increase the current fitness from a value at least~$a$ to a value at least~$b$ is $O(b - a + \log n)$. The number of generations is clearly an upper bound for the number of phases required. 
The previous discussion allows us to apply Wald's equation to conclude that the expected number of evaluations in all phases but the first is bounded by $O(b - a + \log n) \cdot 1/p_{b-1, 1}^+$. Together, this implies the claim.
\end{proof}

We note that for $b-a=\log n$ the bound given by Lemma~\ref{lem:bound-on-expected-evals-to-increase-fitness} depends on the initial offspring population size $\lambda_0$, the gap of fitness to traverse $b-a$ and the probability of finding an improvement at fitness value $b-1$. If we could ensure that $\lambda$ is sufficiently small at the start of the optimisation of every block we could easily compute the total expected optimisation. Unfortunately, the previous lemmas allow for the value of $\lambda$ at the end of a block and hence at the start of a new block to be any large value. We solve this in the following lemma by denoting a generation where $\lambda$ is excessively large as an \emph{excessive} generation. Then, we show that with high probability the algorithm finds the optimum without having an excessive generation. Hence, the expected number of evaluations needed for the algorithm to either find the optimum or have an excessive generation is asymptotically the same as the runtime of the algorithm. 

\begin{lemma}
\label{lem:bound-on-expected-evaluations-or-exceeding-lambda-threshold}
Call a generation~$t$ \emph{excessive} if, for a current search point with fitness~$i$, at the end of the generation $\lambda$ is increased beyond $5F^{1/s}\ln(n)/p_{i, 1}^+$.
Let $T$ denote the expected number of function evaluations before a global optimum is found. Let $\overline{T}$ 
denote the number of evaluations made before a global optimum is found or until the end of the first excessive generation. 
Then
\[
    \E{T} \le \E{\overline{T}} + O(1).
\]
\end{lemma}
\begin{proof}
%
%Define the \oclmax{$g(n)$} as a variant of the \ocl where $\lambda$ is always capped at $g(n)$, for a positive function $g(n) \colon \N \to \N \cup \{\infty\}$. Note that the \oclmax{$\infty$} is identical to the original \ocl and that the \oclmax{$g(n)$} behaves like the \ocl so long as $\lambda$ remains bounded from above by $g(n)$.
%
%Define $\lambda^{(1)} := 5eF^{1/s}n$ and $\lambda^{(2)} := F^{1/s}n^3$. 
%Let $T$ denote the number of function evaluations made by the \ocl before a global optimum is found. Let $T^{(1)}$ 
%denote the number of evaluations made until $\lambda$ increases beyond $\lambda^{(1)}$ or a global optimum is found. 
The proof uses different thresholds for increasingly ``excessive'' values of $\lambda$ and we are numbering the corresponding variables for the number of evaluations and generations, respectively. 
Let $T^{(1)} = \overline{T}$ and let $G^{(1)}$ be the number of \emph{generations} until a global optimum or an excessive generation is encountered. We denote the former event by $B^{(1)}$, that is, the event that an optimum is found before an excessive generation. Let $T^{(2)}$
denote the worst-case number of function evaluations made until $\lambda$ exceeds $\lambda^{(2)} := F^{1/s} n^3$ or the optimum is found, when starting with a worst possible initial fitness and offspring population size $\lambda \le \lambda^{(2)}$. Let $B^{(2)}$ denote the latter event and let $G^{(2)}$ be the corresponding number of generations. Let $T^{(3)}$ denote the worst-case number of evaluations until the optimum is found, when starting with a worst possible population size $\lambda \le \lambda^{(2)}F^{1/s}$. 
Then the expected optimisation time is bounded as follows.
\begin{align*}
    \E{T} \le\;& \E{T^{(1)}}
    + \Prob{\overline{B^{(1)}}} \left(\E{T^{(2)}} + \Prob{\overline{B^{(2)}}} \cdot \E{T^{(3)}}\right).
\end{align*}
Note that $T^{(1)} \le T$ since $T^{(1)}$ is a stopping time defined with additional opportunities for stopping, thus the number of generations $G^{(1)}$ for finding the optimum or exceeding $\lambda^{(1)}$ satisfies $\E{G^{(1)}} = O(n)$ by Lemma~\ref{lem:generations-for-b-a-block}.

% Let $G$ denote the random number of generations until the global optimum is found for the first time, then $\E{G} = O(n)$ by Lemma~\ref{lem:generations-for-b-a-block}. Let $T$ denote the random number of evaluations until the global optimum is found for the first time, then by the law of total probability,
% \begin{align*}
%     \E{T} =\;& \E{T \mid B_{4eF^{1/s}n \ln n}} \Prob{B_{4eF^{1/s}n \ln n}}\\ 
%     & + \E{T \mid B_{n^3F^{1/s}} \cap \overline{B_{4eF^{{1/s}n \ln n}}}} \Prob{B_{n^3 F^{1/s}} \cap \overline{B_{4eF^{{1/s}n \ln n}}}}\\
%     & +  \E{T \mid \overline{B_{n^3F^{1/s}}}} \Prob{\overline{B_{n^3F^{1/s}}}}.
% \end{align*}
In every generation~$t \le G^{(1)}$ with offspring population size $\lambda_t$ and current fitness~$i$, if $\lambda_t \le 5\ln(n)/p_{i, 1}^+$, 
we have $\lambda_{t+1} \le 5F^{1/s}\ln(n)/p_{i, 1}^+$ with probability~1, that is, the generation is not excessive. If $5\ln(n)/p_{i, 1}^+ < \lambda_t \le 5F^{1/s}\ln(n)/p_{i, 1}^+$, we have an excessive generation with probability at most
\[
    (1 - p_{i, 1}^+)^{\lambda_t} \le (1 - p_{i, 1}^+)^{5\ln(n)/p_{i, 1}^+} \le e^{-5\ln(n)} = n^{-5}.
\]
Thus, the probability of having an excessive generation in the first $G^{(1)}$ generations is bounded, using a union bound, by 
\[
    \Prob{\overline{B^{(1)}}} \le \sum_{t=1}^{\infty} t \cdot n^{-5} \cdot \Prob{G^{(1)} = t} = n^{-5} \cdot \E{G^{(1)}} = O(n^{-4}).
\]
%For $\lambda^{(1)} := 5eF^{1/s} n \ln n$, this is at most $\E{G^{(1)}}/n^{-5} = O(n^{-4})$. 
%Defining the number of generations $G^{(2)}$ as the number of generations contributing to $T^{(2)}$, 
We also have $\E{G^{(2)}} = O(n)$ by Lemma~\ref{lem:generations-for-b-a-block} (this bound applies for all initial fitness values and all initial offspring population sizes). In all such generations~$t \le G^{(2)}$, we have $\lambda_t \le \lambda^{(2)}$, thus $\E{T^{(2)}} \le \E{G^{(2)}} \cdot \lambda^{(2)} = O(n^4)$. 

As per the above arguments, 
the probability of exceeding $\lambda^{(2)}$ is either 0 (for $\lambda_t \le \lambda^{(2)}F^{-1/s}$) or (for $\lambda^{(2)}F^{-1/s} < \lambda_t \le \lambda^{(2)}$) bounded by 
\[
    (1 - p_{i, 1}^+)^{\lambda_t} \le (1 - p_{n-1, 1}^+)^{\lambda^{(2)}F^{-1/s}} \le e^{-\Omega(n^2)}.
\]
Taking a union bound,
\[
    \Prob{\overline{B^{(2)}}} \le \sum_{t=1}^{\infty} t \cdot e^{-\Omega(n^2)} \cdot \Prob{G^{(2)} = t} = e^{-\Omega(n^2)} \cdot \E{G^{(2)}} = e^{-\Omega(n^2)}.
\]
%For $\lambda^{(2)} := F^{1/s}n^3$, this is $e^{-\Omega(n^2)} \cdot O(n) = e^{-\Omega(n^2)}$.
Finally, we bound $\E{T^{(3)}} \le n^n$ using the trivial argument that a global optimum is created with every standard bit mutation with probability at least $(1/n)^n$. Putting this together yields
\begin{align*}
    \E{T} \le\;& \E{T^{(1)}}
    + O(n^{-4}) \left(O(n^4) + e^{-\Omega(n^2)} \cdot n^n\right)
    = \E{T^{(1)}} + O(1). 
\end{align*}
\end{proof}

Owing to Lemma~\ref{lem:bound-on-expected-evaluations-or-exceeding-lambda-threshold} we can compute $\E{\overline{T}}$  without worrying about large values of $\lambda$ and at the same time obtain the desired bound on the total expected number of evaluations to find the optimum.

\begin{proof}[Proof of Theorem~\ref{thm:optimisationTime}]
By Lemma~\ref{lem:bound-on-expected-evaluations-or-exceeding-lambda-threshold}, it suffices to bound $\E{\overline{T}}$ from above. In particular, we can assume that no generations are excessive as otherwise we are done. 

We divide the distance to the optimum in \emph{blocks} of length $\log n$ and use this to divide the run into \emph{epochs}. For $i \in \{0, \dots, \lceil n/\log n \rceil-1\}$ Epoch~$i$ starts in the first generation in which the current search point has a fitness of at least $n - (i+1) \log n$ is reached and it ends as soon as a search point of fitness at least $n - i \log n$ is found. Let $T_i$ denote the number of evaluations made during Epoch~$i$. 
Note that after Epoch~$i$, once a fitness of at least $n - i \log n$ has been reached, the algorithm will continue with Epoch~$i-1$ (or an epoch with an even smaller index, in the unlikely event that a whole block is skipped) and the goal of Epoch~0 implies that the global optimum is found. 
Consequently, the total expected number of evaluations is bounded by $\sum_{i=0}^{\lceil n/\log n \rceil -1} \E{T_i}$.

Let $\lambda^{(i)}$ denote the offspring population size at the start of Epoch~$i$.
Applying Lemma~\ref{lem:bound-on-expected-evals-to-increase-fitness} with $a := n - (i+1) \log n$, $b := n - i \log n$ and $\lambda_0 := \lambda^{(i)}$,
\[
    \E{T_i} \le O(\lambda^{(i)}) + O(\log n) \cdot \frac{1}{p_{n - i \log(n) - 1, 1}^+}.
\]
Since we assume that no generation is excessive and the fitness is bounded by $n - i \log(n) -1$ throughout the epoch, we have $\lambda^{(i)} \le 5F^{1/s}\ln(n)/p_{n - i \log(n)-1,1}^+$. Plugging this in, we get
\[
    \E{T_i} \le O(\log n) \cdot \frac{1}{p_{n - i \log(n) - 1, 1}^+} \le O(\log n) \cdot \frac{en}{1 + i \log n} = O(n \log n) \cdot \frac{1}{1 + i \log n}.
\]
Then the expected optimisation time is bounded by 
\begin{align*}
    \sum_{i=0}^{\lceil n/\log n\rceil - 1} \E{T_i}
    \le\;& O(n \log n) \cdot \sum_{i=0}^{\lceil n/\log n\rceil - 1} \frac{1}{1 + i \log n}\\
    \le\;& O(n \log n) \cdot \left(1 + \sum_{i=1}^{\lceil n/\log n\rceil - 1} \frac{1}{i \log n}\right)\\
    =\;& O(n \log n) \cdot \left(1 + \frac{H_{\lceil n/\log n\rceil -1}}{\log n}\right) = O(n \log n)
\end{align*}
using $H_{\lceil n/\log n\rceil -1} \le H_n \le \ln(n) + 1$ in the last step.
\end{proof}

\section{Large Success Rates Fail}\label{sec:exponential-runtime}

In this section, we show that the choice of the success rate is crucial as when $s$ is a large constant,
the runtime becomes exponential.

\begin{theorem}\label{thm:exponential-runtime}
Let the update strength $F \le 1.5$ and the success rate $s\ge 18$ be constants. With probability $1-e^{-\Omega(n/\log^4 n)}$ the \saocl needs at least $e^{\Omega(n/\log^4 n)}$ evaluations to optimise \onemax.
\end{theorem}

The reason why the algorithm takes exponential time is that
now $F^{1/s}$ is small and $\lambda$ only increases slowly in unsuccessful generations, whereas successful generations decrease $\lambda$ by a much larger factor of $F$.
This is detrimental during early parts of the run where it is \emph{easy} to find improvements and there are frequent improvements that decrease~$\lambda$. When $\lambda$ is small, there are frequent fallbacks,
hence the algorithm stays in a region with small values of $\lambda$, where it finds improvements with constant probability, but also has fallbacks with constant probability.
We show, using another potential function based on Definition~\ref{def:potential-function}, that it takes exponential time to escape from this equilibrium.

\begin{definition}\label{def:potential-function-exp}
We define the potential function $g_2(X_t)$ as
\begin{align*}
    g_2(X_t) := f(x_t) + 2.2\log_{F}^2\lambda_t.
\end{align*}
\end{definition}
While $g_1(X_t)$ used a (capped) linear contribution of $\log_F(\lambda_t)$ for $h(\lambda_t)$, here we use the function $h(\lambda_t):= 2.2 \log_F^2(\lambda_t)$ that is convex in $\log_F(\lambda_t)$, so that changes in $\lambda_t$ have a larger impact on the potential.
We show that, in a given fitness interval, the potential $g_2(X_t)$ has a negative drift.

\begin{lemma}\label{lem:potentialDrift_Deltah}
Consider the \saocl as in Theorem~\ref{thm:exponential-runtime}.
Then there is a constant $\delta>0$ such that for every ${0.84 n+2.2\log^2(4.5)< g_2(X_t)< 0.85n}$,
\begin{align*}
    \E{g_2(X_{t+1})-g_2(X_{t})\mid X_{t}} \le -\delta.
\end{align*}
\end{lemma}
\begin{proof}%[Proof of Lemma~\ref{lem:potentialDrift_Deltah}]
We abbreviate ${\Deltah:=\E{g_2(X_{t+1})-g_2(X_{t})\mid X_{t}}}$. Given that for all~$\lambda\ge1$
\begin{align*}
    h(\lambda/F) = 2.2\log_F^2(\lambda/F) = 2.2\left(\log_F(\lambda) - 1\right)^2 \ge 0 = h(1)
\end{align*}
then by Lemma~\ref{lem:generalised-potential}, for all $\lambda$, $\Deltah$ is at most
\begin{align*}
    &\left(\Deltagain+h(\lambda/F)-h(\lambda F^{1/s})\right)\pimp + h(\lambda F^{1/s}) -h(\lambda) -\Deltaloss \ploss\\
    & = \left(\Deltagain +2.2 \log_F^2(\lambda/F) - 2.2 \log_F^2(\lambda F^{1/s})\right)\pimp +\\
    &\hspace{5.3cm} 2.2 \log_F^2(\lambda F^{1/s}) - 2.2 \log_F^2(\lambda) - \Deltaloss\ploss \\
    & = \left(\Deltagain +2.2 (\log_F(\lambda)-1)^2 - 2.2 (\log_F(\lambda)+1/s)^2\right)\pimp +\\
    &\hspace{4.8cm} 2.2 (\log_F(\lambda)+1/s)^2 - 2.2 \log_F^2(\lambda) - \Deltaloss\ploss \\
    & = \left(\Deltagain - \left(1+\frac{1}{s}\right)\cdot 4.4\log_F(\lambda) + 2.2 - \frac{2.2}{s^2}\right)\pimp +\frac{4.4\log_F\lambda}{s}+\frac{2.2}{s^2} - \Deltaloss\ploss.
\end{align*}
The terms containing the success rate $s$ add up to
\begin{align*}
    (1-\pimp)\left(\frac{4.4\log_F\lambda}{s}+\frac{2.2}{s^2}\right).
\end{align*}
This is non-increasing in $s$, thus we bound~$s$ by the assumption  $s\ge18$, obtaining
\begin{align}
    \Deltah \le & \left(\Deltagain - \frac{19}{18}\cdot4.4\log_F\lambda + 2.2 - \frac{2.2}{324}\right)\pimp +\frac{4.4\log_F\lambda}{18}+\frac{2.2}{324} - \Deltaloss\ploss.\label{eq:deltagain}
\end{align}

We note that in Equation~\eqref{eq:deltagain}, $\lambda\in\mathbb{R}_{\ge1}$ but since the algorithm creates $\round{\lambda}$ offspring, the forward drift and the probabilities are calculated using $\round{\lambda}$. In the following in all the computations the last digit is rounded up if the value was positive and down otherwise to ensure the inequalities hold.  We start taking into account only $\round{\lambda}\ge5$, that is, ${\lambda}\ge4.5$ and later on we will deal with smaller values of $\lambda$. With this constraint on $\lambda$ we use the simple bound $\ploss \ge 0$.
Bounding $\Deltagain\le\lceil\log\lambda\rceil+0.413$ using Equation~\eqref{eq:expectedGain} in Lemma~\ref{lem:bounds_probabilites},
\begin{align*}
    \Deltah \le \left(2.613  +\lceil\log\lambda\rceil-\frac{19}{18}\cdot4.4\log_F\lambda\right.
    \left.-\frac{2.2}{324}\right)\pimp + \frac{4.4\log_F\lambda}{18}+\frac{2.2}{324}.
\end{align*}

For all $\lambda\ge1$, ${f(x_t) \ge 0.85n}$ implies $g_2(X_t)\ge 0.85n$. By contraposition, our precondition $g_2(X_t)< 0.85n$ implies ${f(x_t)<0.85n}$. Therefore, using Equation~\eqref{eq:pimp} in Lemma~\ref{lem:bounds_probabilites} with the worst case ${f(x_t)=0.85n}$ and $\round{\lambda}=5$ we get ${\pimp \ge 1-\frac{e}{e+0.15\round{\lambda}}\ge 1-\frac{e}{e+5\cdot0.15}>0.216}$. Substituting these bounds we obtain
\begin{align*}
    \Deltah &\le \left(2.613+\lceil\log\lambda\rceil - \frac{19}{18}\cdot4.4\log_F\lambda-\frac{2.2}{324}\right)0.216 + \frac{4.4\log_F\lambda}{18}+\frac{2.2}{324}\\
    & \le 0.5562 + 0.216\lceil\log\lambda\rceil - 0.7587\log_F\lambda
\end{align*}
The assumption $F\le1.5$ implies that $\log_F\lambda \ge \log \lambda$. Using this and $\lceil\log\lambda\rceil\le\log(\lambda)+1$ yields
% that $\log_F\lambda \ge \log \lambda$ for all $F\le2$ and $\lceil\log\lambda\rceil\le\log(\lambda)+1$,
\begin{align*}
    \Deltah &\le 0.5562 + 0.216(\log(\lambda)+1) - 0.7587\log\lambda\\
    & = 0.7722 - 0.5427\log\lambda\\
    & \le 0.7722 - 0.5427\log 4.5 \le -0.4054.
\end{align*}
Up until now we have proved that $\Deltah\le -0.4058$ for all ${f(x_t)<0.85n}$ and $\round{\lambda}\ge5$. Now we need to consider ${\round{\lambda}<5}$.
For $\round{\lambda} < 5$, that is, $\lambda < 4.5$, the precondition ${g_2(X_t) > 0.84 n+2.2\log^2(4.5)}$ implies that $f(x_t) > 0.84n$.
Therefore, the last part of this proof focuses only on $0.84n<f(x_t)<0.85n$ and $\round{\lambda}<5$. For this region we use again Equation~\eqref{eq:deltagain}, but bound it in a more careful way now. By Equation~\eqref{eq:ploss} in Lemma~\ref{lem:bounds_probabilites}, $\ploss\ge\left(\frac{f(x_t)}{n}-\frac{1}{e}\right)^{\round{\lambda}}\ge \left(0.84-\frac{1}{e}\right)^{\round{\lambda}}$ and bounding $\Deltagain$ and $\Deltaloss$ using Equations~\eqref{eq:expectedGain} and \eqref{eq:expectedLoss} in Lemma~\ref{lem:bounds_probabilites} yields:
\begin{multline}\label{eq:exactdeltah}
    \Deltah \le \underbrace{\left(\sum_{j=1}^\infty \left(1-\left(1-\frac{1}{j!}\right)^{\round{\lambda}}\right)- \frac{19}{18}\cdot4.4\log_F\lambda + 2.2 - \frac{2.2}{324}\right)}_{\alpha} \pimp\\
    + \frac{4.4\log_F\lambda}{18}+\frac{2.2}{324} - \left(0.84-\frac{1}{e}\right)^{\round{\lambda}}.
\end{multline}
We did not bound $\pimp$ in the first term yet because the factor $\alpha$ in brackets preceding it can be positive or negative. We now calculate precise values for $\sum_{j=1}^\infty\left(1-\left(1-\frac{1}{j!}\right)^{\round{\lambda}}\right)$ giving $e-1$, $2.157$, $2.4458$ and $2.6511$ for $\round{\lambda}=1, 2, 3, 4$, respectively. Given that $F\le 1.5$ the factor $\alpha$ is negative for all $1.5\le \lambda<4.5$, because
\begin{align*}
    &\left(\sum_{j=1}^\infty \left(1-\left(1-\frac{1}{j!}\right)^{\round{\lambda}}\right)- \frac{19}{18}\cdot4.4\log_F\lambda + 2.2 - \frac{2.2}{324}\right) \\
    &\le\left(\sum_{j=1}^\infty \left(1-\left(1-\frac{1}{j!}\right)^{\round{\lambda}}\right)- 4.4\log_F(\lambda) + 2.2\right) \\
    &\le
    \begin{cases}
    4.8511-4.4\log_{1.5}(3.5)= -8.74 & 3.5\le\lambda<4.5\\
    4.6458-4.4\log_{1.5}(2.5)= -5.29 & 2.5\le\lambda<3.5\\
    4.357-4.4\log_{1.5}(1.5)= -0.043 & 1.5\le\lambda<2.5
    \end{cases}
\end{align*}

On the other hand, for $\lambda<1.5$ and $\round{\lambda}=1$, $\alpha$ is positive when $\lambda < F^{\gamma}$ for $\gamma=\frac{1933}{7524} + \frac{45 e}{209}\approx 0.8422$ and negative otherwise. With this we evaluate different ranges of $\lambda$ separately using Equation~\eqref{eq:exactdeltah}.
For $1 \le \lambda < F^{\gamma}$, we get $\round{\lambda}=1$ and by Lemma~\ref{lem:bounds_probabilites} if $0.84n\le i\le0.85n$ and $n\ge163$ then $\pimp\le 0.069$, thus
\begin{align*}
    \Deltah & \le \left(e+1.2-\frac{19}{18} \cdot 4.4\log_F(\lambda)- \frac{2.2}{324}\right) 0.069 + \frac{4.4}{18}\log_F(\lambda) +
    \frac{2.2}{324} - \left(0.84-\frac{1}{e}\right)\\
    & \le -0.076\log_F(\lambda) -0.195 \le -0.195.
\end{align*}
For $F^{\gamma} \le \lambda < 1.5$, by Lemma~\ref{lem:bounds_probabilites} we bound $\pimp\ge\frac{n-f(x_t)}{en} \ge 0.0551$:
\begin{align*}
    \Deltah & \le \left(e+1.2-\frac{19}{18} \cdot 4.4\log_F(\lambda)- \frac{2.2}{324}\right) 0.0551+ \frac{4.4}{18}\log_F(\lambda) + \nonumber\\
    &\hspace{7cm}\frac{2.2}{324} - \left(0.84-\frac{1}{e}\right)\\
    & \le -0.0114 \log_F (\lambda) - 0.2498 \le - 0.2498.
\end{align*}
For $1.5 \le \lambda < 2.5$, by Lemma~\ref{lem:bounds_probabilites} we bound $\pimp\ge1-\frac{e}{e+0.3} \ge 0.0993$
\begin{align*}
    \Deltah &\le \left(4.357 - \frac{19}{18} \cdot 4.4\log_F(\lambda) -\frac{2.2}{324} \right)0.0993 +\frac{4.4}{18}\log_F(\lambda)+\nonumber\\
    &\hspace{7cm}\frac{2.2}{324} - \left(0.84-\frac{1}{e}\right)^2\\
    & \le 0.2159 -0.2167 \log_F (\lambda)\\
    &\le 0.2159 -0.2167 \log_{1.5} (1.5) \le -0.0008.
\end{align*}
For $2.5\le \lambda <3.5$ we use $\pimp\ge1-\frac{e}{e+0.45}=0.142$,
\begin{align*}
    \Deltah &\le \left(4.6458 - \frac{19}{18} \cdot 4.4\log_F(\lambda) -\frac{2.2}{324} \right)0.142 +\frac{4.4}{18}\log_F(\lambda)+\nonumber\\
    &\hspace{7cm}\frac{2.2}{324} - \left(0.84-\frac{1}{e}\right)^3\\
    & \le 0.5612 - 0.415 \log_F (\lambda)\\
    & \le 0.5612 - 0.415 \log_{1.5} (2.5) \le -0.376.
\end{align*}
Finally for $3.5\le \lambda <4.5$ we use $\pimp\ge1-\frac{e}{e+0.6}=0.1808$, \begin{align*}
    \Deltah &\le \left(4.8511 - \frac{19}{18} \cdot 4.4\log_F(\lambda) -\frac{2.2}{324} \right)0.1808 +\frac{4.4}{18}\log_F(\lambda)+\nonumber\\
    &\hspace{7cm}\frac{2.2}{324} - \left(0.84-\frac{1}{e}\right)^4\\
    & \le 0.832 - 0.5952 \log_F (\lambda) \\
    & \le 0.832 - 0.5952 \log_{1.5} (3.5) \le -1.006.
\end{align*}

With these results we can see that the potential is negative with $\lambda\in [1,4.5)$ and $0.84n<f(x_t)<0.85n$. Hence, for every $0.84 n+\log^2(4.5)< g_2(X_t)< 0.85n$, and $\delta = 0.0008$, $\Deltah \le -\delta$.
\end{proof}

\begin{figure}[h]
\centering
\includegraphics[width=\linewidth]{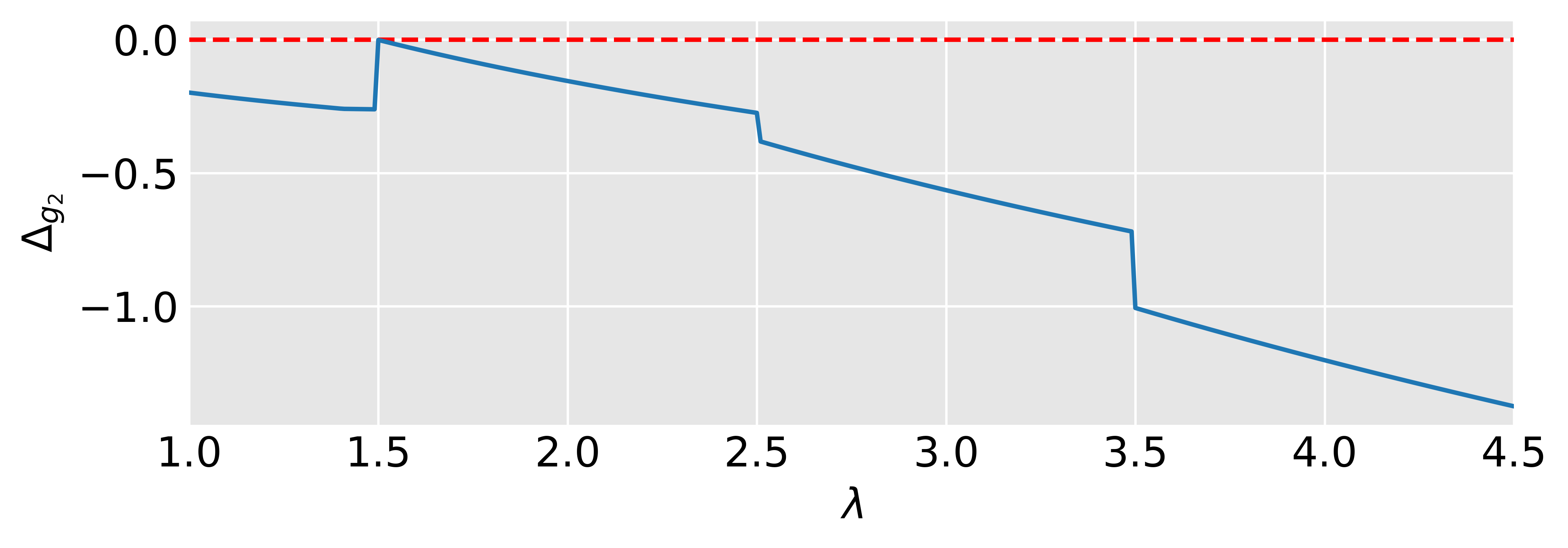}
\caption{Bounds on $\Delta_{g_2}$ with a maximum of $-0.0008$ for $\lambda=1.5$.}
\label{fig:drift-deltah-bound}
\end{figure}

Finally, with Lemmas \ref{lem:potentialDrift_Deltah} and \ref{lem:probability-exceed-polyn}, we now prove Theorem \ref{thm:exponential-runtime}.

\begin{proof}[Proof of Theorem \ref{thm:exponential-runtime}]
We apply the negative drift theorem with scaling (Theorem~\ref{thm:negative-drift-with-scaling}). We switch to the potential function $\overline{h}(X_t):=\max\{0, n-g_2(X_t)\}$ in order to fit the perspective of the negative drift theorem. In this case we can pessimistically assume that if $\overline{h}(X_t) = 0$ the optimum has been found.

The first condition of the negative drift theorem with scaling (Theorem~\ref{thm:negative-drift-with-scaling}) can be established with Lemma \ref{lem:potentialDrift_Deltah} for $a=0.15n$ and $b=0.16n-2.2\log^2(4.5)$. Furthermore, with Chernoff bounds we can prove that at initialization $\overline{h}(X_t)\ge b$ with probability $1 - 2^{-\Omega(n)}$.

To prove the second condition we need to show that the probability of large jumps is small. Starting with the contribution that $\lambda$ makes to the change in $\overline{h}(X_t)$, we use Lemma \ref{lem:probability-exceed-polyn} to show that this contribution is at most ${2.2 \log^2 (e F^{1/s} n^3) \le 4.79 + 19.8 \log^2 n\le 20 \log^2 n}$ with probability $\exp(-\Omega(n^2))$, where the last inequality holds for large enough~$n$.

The only other contributor is the change in fitness. The probability of a jump in fitness away from the optimum is maximised when there is only one offspring. On the other hand the bigger the offspring population the higher the probability of a large jump towards the optimum. Taking this into account and pessimistically assuming that every bit flip either decreases the fitness in the first case or increases the fitness in the latter we get the following probabilities. Recalling~\eqref{eq:distribution-of-flipping-bits},
\begin{align*}
    \prob{f(x_t)-f(x_{t+1})\ge \kappa}\le\frac{1}{\kappa!}
\end{align*}
\begin{align*}
    \prob{f(x_{t+1})-f(x_{t})\ge \kappa}\le1-\left(1-\frac{1}{\kappa!}\right)^\lambda
\end{align*}
Given that $\frac{1}{\kappa!}\le 1-\left(1-\frac{1}{\kappa!}\right)^\lambda$, and that $\lambda\le e F^{1/s} n^3$.
\begin{align*}
    \prob{\vert f(x_{t+1})-f(x_{t})\vert\ge \kappa}
    &\le 1-\left(1-\frac{1}{\kappa!}\right)^{e F^{1/s} n^3}\\
    &\le \frac{e F^{1/s} n^3}{\kappa!}\\
    &\le \frac{e^{\kappa+1} F^{1/s} n^3}{\kappa^\kappa}
\end{align*}
Joining both contributions, we get
\begin{align}\label{eq:prob_large_jumps}
    \prob{\vert g_2(X_{t+1})-g_2(X_{t})\vert\ge \kappa + 20 \log^2 n} \le \frac{e^{\kappa+1} F^{1/s} n^3}{\kappa^\kappa}.
\end{align}
To satisfy the second condition of the negative drift theorem with scaling (Theorem~\ref{thm:negative-drift-with-scaling}) we use $r=21 \log^2 n$ and ${\kappa = j \log^2 n}$ in order to have $\kappa + 20 \log^2 n\le jr$ for $j\in \N$. For $j=0$ the condition $\prob{\vert g_2(X_{t+1})-g_2(X_{t})\vert\ge jr} \le e^{0}$ is trivial. From Equation~\eqref{eq:prob_large_jumps}, we obtain
\begin{align*}
    \prob{\vert g_2(X_{t+1})-g_2(X_{t})\vert\ge jr}
    &\le \frac{e F^{1/s} e^{(j \log^2 n)} n^3}{(j \log^2 n)^{j \log^2 n}}
\end{align*}
We simplify the numerator using
\[
    e^{(j \log^2 n)} n^3 = e^{(j \log(n)\ln(n)/\ln(2))} n^3 = n^{(3 + j \log(n)/\ln(2))}
\]
and bound the denominator as
\begin{align*}
& (j \log^2 n)^{j \log^2 n} \ge
(\log n)^{2j \log^2 n}
= n^{2j \log(n)\log \log(n)},
\end{align*}
yielding
\begin{align*}
    \prob{\vert g_2(X_{t+1})-g_2(X_{t})\vert\ge jr}
    \le\;& e F^{1/s} n^{(3+j \log(n)/\ln(2)-2 j (\log n)  \log \log n)}\\
    =\;& e F^{1/s} n^{(3+j (\log(n)/\ln(2)-2 (\log n)  \log \log n))}.
\end{align*}
For $n\ge7$, $\log(n)/\ln(2)-2 (\log n)  \log \log n \le -4$, hence
\begin{align*}
    \prob{\vert g_2(X_{t+1})-g_2(X_{t})\vert\ge jr}
    \le\;& e F^{1/s} n^{3-4j}.
\end{align*}
which for large enough~$n$ is bounded by $e^{-j}$ as desired. 

The third condition is met with $r=21 \log^2 n$ given that $\delta \ell / (132 \log((21 \log^2 n)/\delta)) = \Theta (n/\log\log n)$, which is larger than $r^2 = \Theta(\log^4 n)$ for large enough~$n$.

With this we have proved that the algorithm needs at least $e^{\Omega(n/\log^4 n)}$ generations with probability $1-e^{-\Omega(n/\log^4 n)}$. Since each generation uses at least one fitness evaluation, the claim is proved.
\end{proof}

We note that although Theorem~\ref{thm:exponential-runtime} is applied for \onemax specifically, the conditions used in the proof of Theorem~\ref{thm:exponential-runtime} and Lemma~\ref{lem:potentialDrift_Deltah} apply for several other benchmark functions. This is because our result only depends on some fitness levels of \onemax and other functions have fitness levels that are symmetrical or resemble these fitness levels. We show this in the following theorem. To improve readability we use $\ones{x}:= \sum_{i=1}^{n} x_i$ and ${\zeros{x}:= \sum_{i=1}^{n} (1-x_i)}$.

\begin{theorem}\label{thm:extended_functions}
Let the update strength $F \le 1.5$ and the success rate $s\ge 18$ be constants. With probability $1-e^{-\Omega(n/\log^4 n)}$ the \saocl needs at least $e^{\Omega(n/\log^4 n)}$ evaluations to optimise:
% \jumpk with $k=o(n)$, \cliffd with $d=o(n)$, \zeromax, \twomax and \ridge. These functions are defined as:
\begin{itemize}[label=\textbullet]
    \item $\textsc{Jump}_k(x) :=
    \begin{cases}
        n - \ones{x} & \textrm{if}\ n-k<\ones{x}<n, \\
        k + \ones{x} & \textrm{otherwise},
    \end{cases}$
    \\ with $k=o(n)$,
    \item $\cliffd(x) :=
    \begin{cases}
        \ones{x} & \textrm{if}\ \ones{x} \leq d, \\
        \ones{x} -d + 1/2 & \textrm{otherwise},
    \end{cases}$
    \\ with $d=o(n)$,
    \item $\zeromax(x):= \zeros{x}$,
    \item $\twomax(x):=\max\left\{\ones{x}, \zeros{x}\right\}$,
    \item $\ridge(x):=
    \begin{cases}
        n+\ones{x} & \textrm{if}\ x=1^i0^{n-i}, i\in\{0,1,\dots,n\}, \\
        \zeros{x} & \textrm{otherwise}.
    \end{cases}$
\end{itemize}
\end{theorem}
\begin{proof}
For \jumpk and \cliffd, given that $k$ and $d$ are $o(n)$ the algorithm needs to optimise a \onemax-like slope with the same transition probabilities as in Lemma~\ref{lem:potentialDrift_Deltah} before the algorithm reaches the local optima.
% before the algorithm reaches the local optima it needs to optimise a \onemax-like slope with the same transition probabilities as in Lemma~\ref{lem:potentialDrift_Deltah},
Hence, we can apply the negative drift theorem with scaling (Theorem~\ref{thm:negative-drift-with-scaling}) as in Theorem~\ref{thm:exponential-runtime} to prove the statement.

For \zeromax the algorithm will behave exactly as in \onemax, because it is unbiased towards bit-values. Similarly, for \twomax, independently of the slope the algorithm is optimising, it needs to traverse through a \onemax-like slope needing at least the same number of function evaluations as in \onemax.

Finally, for \ridge, unless the algorithm finds a search point on the ridge (${x \in 1^i0^{n-i}}$ with $i\in\{0,1,\dots,n\}$) beforehand, the first part of the optimisation behaves as \zeromax and similar to Theorem~\ref{thm:exponential-runtime} by Lemma~\ref{lem:potentialDrift_Deltah} and the negative drift theorem with scaling (Theorem~\ref{thm:negative-drift-with-scaling}) it will need at least $e^{C n/\log^4 n}$ generations with probability $e^{-C n/\log^4 n}$ to reach a point with $\ones{x}\le 0.15n$ for some constant $C>0$. 

It remains to show that the ridge is not reached during this time, with high probability.
We first imagine the algorithm optimising \zeromax and note that the behaviour on \ridge and \zeromax is identical as long as no point on the ridge is discovered. Let $x_0, x_1, \dots$ be the search points created by the algorithm on \zeromax in order of creation. Since \zeromax is symmetric with respect to bit positions, for any arbitrary but fixed~$t$ we may assume that the search point $x_t$ with $d=\ones{x_t}$ is chosen uniformly at random from the $\binom{n}{d}$ search points that have exactly $d$ $1$-bits. There is only one search point $1^d 0^{n-d}$ that on the function \ridge would be part of the ridge. Thus, for $d\ge0.15n$ the probability that $x_t$ lies on the ridge is at most
\begin{align*}
    \binom{n}{d}^{-1}\le\binom{n}{0.15n}^{-1}\le \left(\frac{n}{0.15n}\right)^{-0.15n}= \left(\frac{20}{3}\right)^{-0.15n}.
\end{align*}
(Note that these events for $t$ and $t'$ are not independent; we will resort to a union bound to deal with such dependencies.)
By Lemma~\ref{lem:probability-exceed-polyn}, during the optimisation of any unimodal function every generation uses $\lambda\le eF^{1/s}n^3$ with probability $1-\exp{(-\Omega(n^2))}$. By a union bound over $e^{C n/\log^4 n}$ generations, for an arbitrary constant~$C > 0$, each generation creating at most $eF^{1/s}n^3$ offspring, the probability that a point on the ridge is reached during this time is at most
\begin{align*}
    e^{Cn/\log^4 n} \cdot eF^{1/s}n^3 \cdot \left(\frac{20}{3}\right)^{-0.15n} = e^{-\Omega(n)}.
\end{align*}
Adding up all failure probabilities, the algorithm will not create a point on the ridge before $e^{C n/\log^4 n}$ generations have passed with probability ${1-e^{-\Omega(n/\log^4 n)}}$, and the algorithm needs at least $e^{\Omega(n/\log^4 n)}$ evaluations to solve \ridge with probability $1-e^{-\Omega(n/\log^4 n)}$.
\end{proof}

\section{Experiments}
\label{sec:experiments}

Due to the complex nature of our analyses there are still open questions about the behaviour of the algorithm. In this section we show some elementary experiments to enhance our understanding of the parameter control mechanism and address these unknowns. All the experiments were performed using the IOHProfiler~\cite{IOHProfiler}.

In Section~\ref{sec:poly-runtime} we have shown that both the \saocl and the self-adjusting $(1+\lambda)$~EA have an asymptotic runtime of $O(n\log n)$ evaluations on \onemax. This is the same asymptotic runtime as the \ocl with static parameters $\lambda=\left\lceil\log_{\frac{e}{e-1}}(n)\right\rceil$ \cite{Rowe2014}. 
We remark that very recently the conditions for efficient offspring population sizes have been relaxed to $\lambda \ge \left\lceil\log_{\frac{e}{e-1}}(cn/\lambda)\right\rceil$ for any constant $c > e^2$~\cite{Bossek2021a}. However, this only reduces the best known value of $\lambda$ by 1 or 2 for the considered problem sizes, and so we stick to the simpler formula of $\lambda=\left\lceil\log_{\frac{e}{e-1}}(n)\right\rceil$, i.\,e.\ the best static parameter value reported in~\cite{Rowe2014}.
Unfortunately the asymptotic notation may hide large constants, therefore, our first experiments focus on the comparison of these three algorithms on \onemax.

\begin{figure}[h!]
\centering
\includegraphics[width=0.8\linewidth]{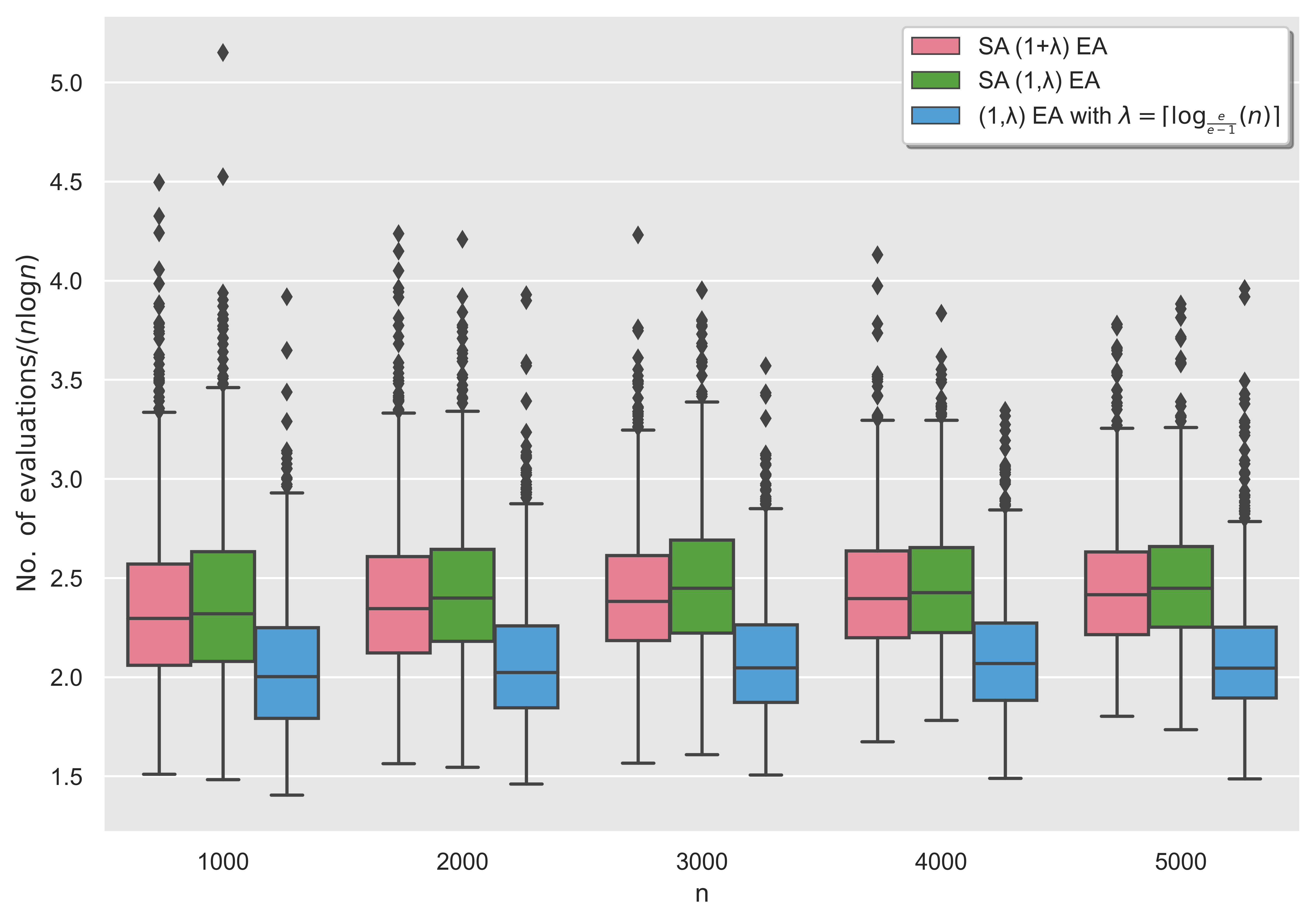}
\caption{Box plots of the number of evaluations used by the \saocl, the self-adjusting $(1+\lambda)$~EA with $s=1$, $F=1.5$ and the \ocl  over $1000$ runs for different~$n$ on \onemax. The number of evaluations is normalised by $n \log n$.}
\label{fig:comparison-elitist-non-elitist}
\end{figure}

Figure~\ref{fig:comparison-elitist-non-elitist} displays box plots of the number of evaluations over 1000 runs for different problem sizes on \onemax.
From Figure~\ref{fig:comparison-elitist-non-elitist} we observe that the difference between both self-adjusting algorithms is relatively small. This indicates that there are only a small number of fallbacks in fitness and such fallbacks are also small.
We also observe that the best static parameter choice from~\cite{Rowe2014} is only a small constant factor faster than the self-adjusting algorithms.
%; in particular for \leadingones we found that for $n\ge2000$ both self-adjusting algorithms follow the same evolutionary path in all runs when using the same random seed generator. That is, there are no fallbacks in fitness for the non-elitist algorithm. This last observation was expected from our analysis, because $\lambda$ grows to at least a sublinear value before finding an improvement and stays large during the optimisation behaving as an elitist algorithm.

\begin{figure}[h!]
\centering
\includegraphics[width=\linewidth]{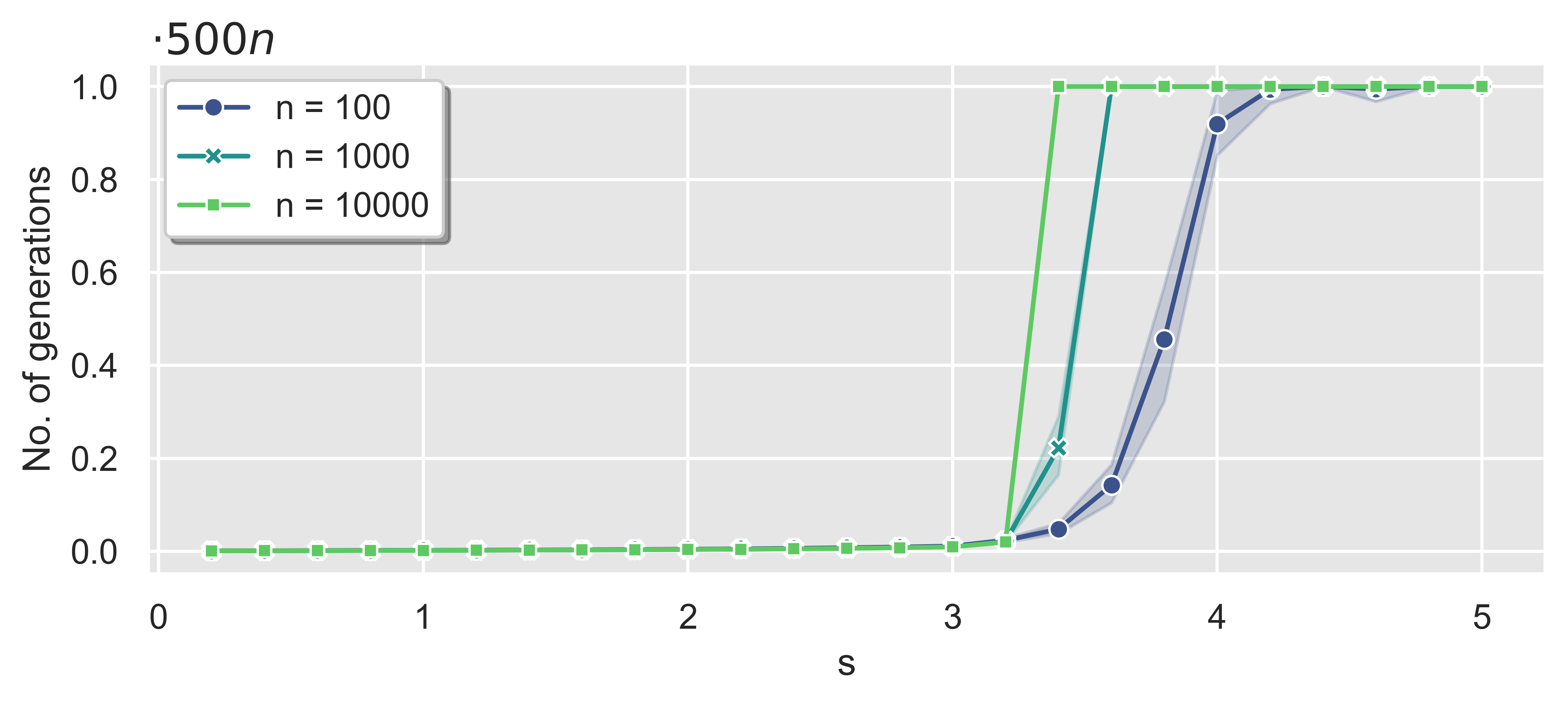}
\caption{Average number of generations with 99\% bootstrapped confidence interval of the \saocl with $F=1.5$ in $100$ runs for different~$n$, normalised and capped at $500n$ generations.}
\label{fig:empirical-tests}
\end{figure}

In the results of Sections~\ref{sec:poly-runtime} and~\ref{sec:exponential-runtime} there is a gap between $s < 1$ and $s \ge 18$ where we do not know how the algorithm behaves on \onemax. In our second experiment, we explore how the algorithm behaves in this region by running the \saocl on \onemax using different values for $s$ shown in Figure~\ref{fig:empirical-tests}. All runs were stopped once the optimum was found or after $500n$ generations were reached. We found a sharp threshold at $s\approx3.4$, indicating that the widely used one-fifth rule ($s=4$) is inefficient here, but other success rules achieve optimal asymptotic runtime.

\begin{figure}[h!]
\centering
\includegraphics[width=\linewidth]{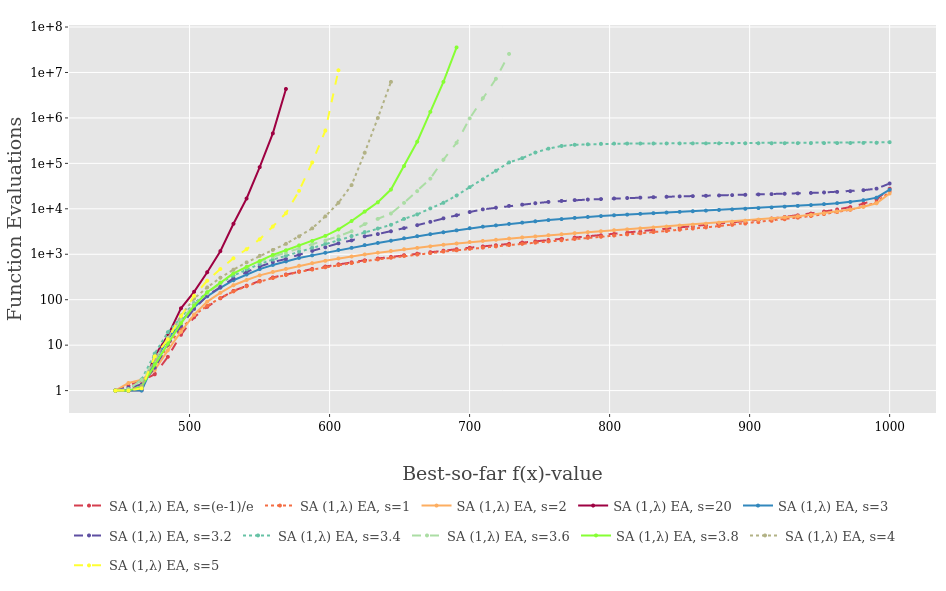}
\caption{Fixed target results for the \saocl on \onemax with $n=1000$ (100 runs).}
\label{fig:fixed_target}
\end{figure}

Additionally, in Figure~\ref{fig:fixed_target} we plot fixed target results, that is, the average time to reach a certain fitness, for $n=1000$ for different $s$. All runs were stopped once the optimum was found or after $500n$ generations. No points are graphed for fitness values that were not reached during the allocated time. We note that the plots do not start exactly at $n/2=500$; this is due to the random effects of initialisation. From here we found that the range of fitness values with negative drift is wider than what we where able to prove in Section~\ref{sec:exponential-runtime}. Already for $s=3.4$, there is an interval on the scale of number of ones around $0.7n$ where the algorithm spends a large amount of evaluations to traverse this interval. Interestingly, as $s$ increases the algorithm takes longer to reach points farther away from the optimum.

\begin{figure}[h!]
\centering
\includegraphics[width=\linewidth]{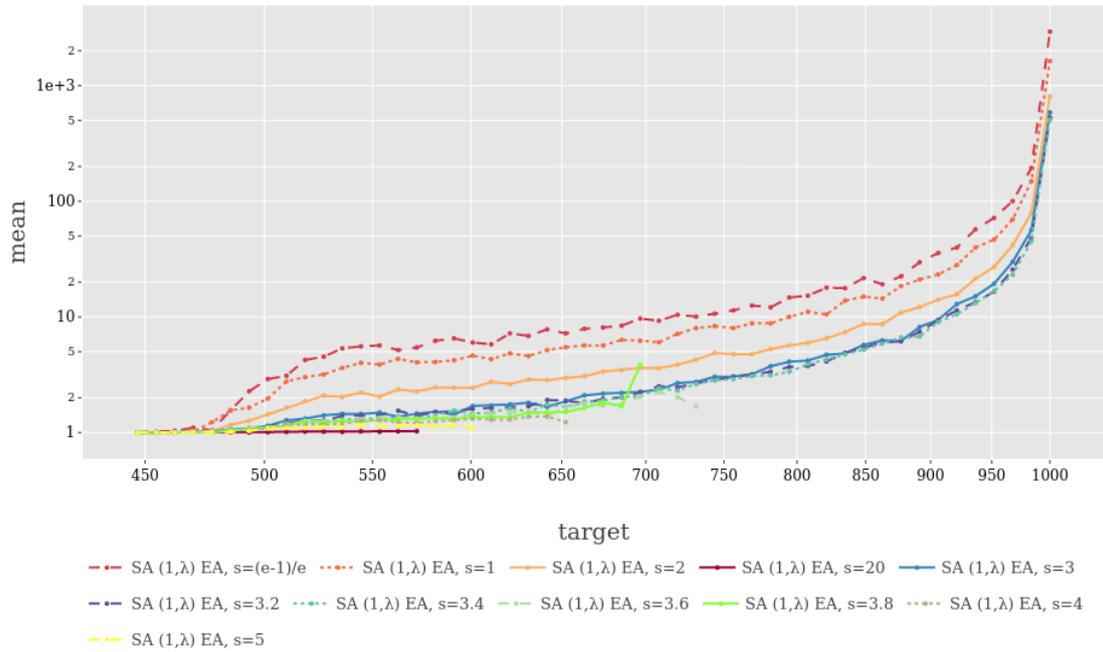}
\caption{Average $\lambda$ values for each fitness level of the \saocl on \onemax with $n=1000$ (100 runs).}
\label{fig:average_lambda_per_fitness}
\end{figure}

We also explored how the parameter $\lambda$ behaves throughout the optimisation depending on the value of $s$. In Figure~\ref{fig:average_lambda_per_fitness} we can see the average $\lambda$ at every fitness value for $n=1000$. As expected, on average $\lambda$ is larger when $s$ is smaller. For $s\ge 3$ we can appreciate that on average $\lambda<2$ until fitness values around $0.7n$ are reached. This behaviour is what creates the non-stable equilibrium slowing down the algorithm.

\begin{figure}[h!]
\centering
\includegraphics[width=\linewidth]{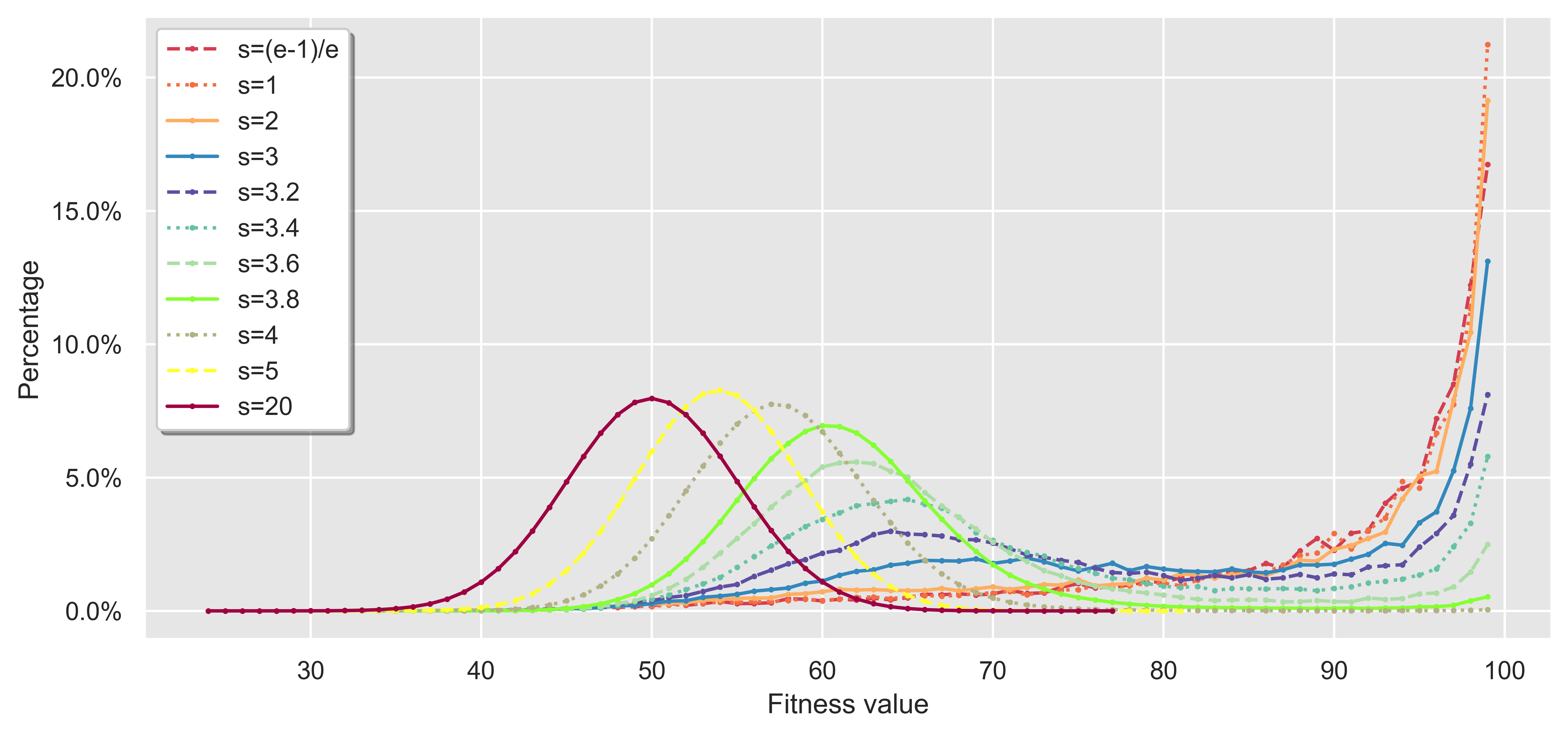}
\caption{Percentage of fitness function evaluations used per fitness value for the \saocl on \onemax with $n=100$ over 100 runs (runs were stopped when the optimum was found or when 1,500,000 function evaluations were made).}
\label{fig:evaluations_per_fitness}
\end{figure}

Finally, to identify the area of attraction of the non-stable equilibrium, 
in Figure~\ref{fig:evaluations_per_fitness} we show the percentage of fitness evaluations spent in each fitness level for $n=100$ (100 runs) and different $s$ values near the transition between polynomial and exponential.
% we implemented a set of experiments (100 runs) with $n=100$ and plotted (Figure~\ref{fig:evaluations_per_fitness}) the percentage of time spent in each fitness level for different $s$ values near the transition between polynomial and exponential.
Runs were stopped when the optimum was found or when 1,500,000 function evaluations were made. The first thing to notice is that for $s=20$ the algorithm is attracted and spends most of the time near $n/2$ ones, which suggests that it behaves similar to a random walk. When $s$ decreases, the area of attraction moves towards the optimum but stays at a linear distance from it.
%and even for $s=3.2$ there is a small area of attraction that causes a ``hump'' in the graph around $0.65n$. 
For $s \le 3.4$ most of the evaluations are spent near the optimum on the harder fitness levels where $\lambda$ tends to have linear values.

\section{Discussion and Conclusions}

We have shown that simple success-based rules, embedded in a \ocl, are able to optimise \onemax in $O(n)$ generations and $O(n \log n)$ evaluations. The latter is best possible for any unary unbiased black-box algorithm~\cite{Lehre2012,Doerr20201}.

However, this result depends crucially on the correct selection of the success rate $s$. The above holds for constant $0 < s < 1$ and, in sharp contrast, the runtime on \onemax (and other common benchmark problems) becomes exponential with overwhelming probability if $s \ge 18$.
Then the algorithm stagnates in an equilibrium state at a linear distance to the optimum where successes are common. Simulations showed that, once $\lambda$ grows large enough to escape from the equilibrium, the algorithm is able to maintain large values of $\lambda$ until the optimum is found. Hence, we observe the counterintuitive effect that for too large values of~$s$, optimisation is harder when the algorithm is far away from the optimum and becomes easier when approaching the optimum. (To our knowledge, such an effect was only reported before on \textsc{HotTopic} functions~\cite{Lengler2021} and Dynamic \textsc{BinVal} functions~\cite{Johannes2021}.)

There is a gap between the conditions $s < 1$ and $s \ge 18$. Further work is needed to close this gap.
In our experiments we found a sharp threshold at $s\approx3.4$, indicating that the widely used one-fifth rule ($s=4$) is inefficient here, but other success rules achieve optimal asymptotic runtime.

Our analyses focus mostly on \onemax, but we also showed that when $s$ is large the \saocl has an exponential runtime with overwhelming probability on \jumpk, \cliffd, \zeromax, \twomax and \ridge. We believe that these results can be extended for many other functions: we conjecture that for any function that has a large number of contiguous fitness levels that are easy, that is, that the probability of a successful generation with $\lambda=1$ is constant, then there is a (large) constant success rate $s$ for which the \saocl would have an exponential runtime. We suspect that many combinatorial problem instances are easy somewhere, for example problems like minimum spanning trees, graph colouring, \textsc{Knapsack} and \textsc{MaxSat} tend to be easy in the beginning of the optimisation. Furthermore, given that for large values of~$s$ the algorithm gets stuck on easy parts of the optimisation and that \onemax is the easiest function with a unique optimum for the \oea~\cite{Doerr2010,Witt13,Sudholt13}, we conjecture that any $s$ that is efficient on \onemax would also be a good choice for any other problem.
%, that is, any $s$ that is efficient on \onemax has at least a similar performance on any other problem than any other constant $s$ on the same problem.

\section*{Declarations}

\subsection*{Compliance with Ethical Standards}
The authors have no competing interests to declare that are relevant to the content of this article.

\subsection*{Funding}
This research has been supported by CONACYT (Consejo Nacional de Ciencia y Tecnolog\'ia) under the grant no. 739621. 

\subsection*{Acknowledgements}
The authors thank an anonymous reviewer as well as the authors of~\cite{KaufmannArxiv2022,KaufmannPPSN2022} for reporting a bug in~\cite{Hevia2021} and a previous version of this manuscript, and Maxime Larcher and Johannes Lengler in particular for discussions on how to fix this bug.

\bibliographystyle{plain}
\bibliography{arXiv2}

\appendix

\section{Proof of Lemma~\ref{lem:global-failures}}

\printProofs
\end{document}